\documentclass[twoside,9pt]{article}

\usepackage[abbrvbib, preprint]{jmlr2e}

\usepackage{multirow}
\usepackage{tabularx}
\usepackage{booktabs}
\usepackage{mathrsfs}
\usepackage[skip=2pt,font=scriptsize, justification=centering]{caption}
\usepackage{bm}
\usepackage{amsmath}
\usepackage{nccmath}
\usepackage{ amssymb }
\usepackage{dsfont}
\usepackage{media9}
\usepackage{graphicx}
\usepackage{MnSymbol,wasysym}
\usepackage{subcaption}
\usepackage{textcomp}
\usepackage{wrapfig}
\usepackage{enumitem}
\usepackage{fancyhdr}
\usepackage{hyperref}
\usepackage{xcolor}
\usepackage{float}
\usepackage{makecell}
\usepackage[title]{appendix}
\usepackage[ruled,vlined]{algorithm2e}

\newtheorem{defi}{Definition}[section]

\newtheorem{prop}{Proposition}[section]

\newcommand{\bP}{\mathbb{P}}
\newcommand{\bE}{\mathbb{E}}
\newcommand{\bV}{\mathbb{V}}

\newcommand{\cD}{\mathcal{D}}

\newcommand{\bM}{\bm{m}}
\newcommand{\bC}{\bm{C}}
\newcommand{\bW}{\bm{W}}
\newcommand{\bG}{\mathbf{G}}
\newcommand{\bias}{\mathbf{b}}
\newcommand{\bu}{\bm{u}}
\newcommand{\cN}{\mathcal{N}}
\newcommand{\rN}{\mathscr{N}}
\newcommand{\btheta}{\boldsymbol{\theta}}
\newcommand{\bTheta}{\bm{\Theta}}
\newcommand{\cG}{\bm{\mathcal{G}}}

\DeclareMathOperator*{\argmax}{arg\,max}

\jmlrheading{1}{2000}{1-48}{4/00}{10/00}{##############}{Ziyang Ding and Sayan Mukherjee}

\ShortHeadings{Intersection of Deep Sequential Model and State-space Model for Option Pricing}{Ding and Mukherjee}
\firstpageno{1}

\begin{document}

\title{At the Intersection of Deep Sequential Model Framework and State-space Model Framework: \\
        Case Study on Option Pricing}

\author{\name Ziyang Ding \email zd26@stanford.edu \\
        \addr Departments of Statistics\\
        Stanford University\\
        \AND
        \name Sayan Mukherjee \email sayan@stat.duke.edu \\
        \addr Departments of Statistical Science \\
        Mathematics, and Computer Science \\
        Duke University \\
        Durham, NC 27708, USA
}

\editor{Editor Happy and Editor Thanksgiving}

\maketitle

\begin{abstract}%   <- trailing '%' for backward compatibility of .sty file
Inference and forecast problems of the nonlinear dynamical system have arisen in a variety of contexts. Reservoir computing and deep sequential models, on the one hand, have demonstrated efficient, robust, and superior performance in modeling simple and chaotic dynamical systems. However, their innate deterministic feature has partially detracted their robustness to noisy system, and their inability to offer uncertainty measurement has also been an insufficiency of the framework. On the other hand, the traditional state-space model framework is robust to noise. It also carries measured uncertainty, forming a just-right complement to the reservoir computing and deep sequential model framework. We propose the unscented reservoir smoother, a model that unifies both deep sequential and state-space models to achieve both frameworks' superiorities. Evaluated in the option pricing setting on top of noisy datasets, URS strikes highly competitive forecasting accuracy, especially those of longer-term, and uncertainty measurement. Further extensions and implications on URS are also discussed to generalize a full integration of both frameworks. 

\end{abstract}


\section{Introduction} \label{sec:Introduction}

%######### RC

\textbf{\textit{The reservoir computer (RC)}}  (\citet{Jaeger2002}, \citet{Jaeger2004}, \citet{Maass2000}, \citet{Maass2002}, \citet{Crook2007}, \citet{Verstraeten2007})is a family of RNN models whose recurrent part is kept fixed and is either generated randomly, or by custom topologies to facilitate the information flow (\citet{Jaeger2007},  \citet{Lukosevicius2009}, \citep{Scardapane2017}, \citep{Rodan2017}). Its recurrence structure is determined by two mappings, namely a reservoir evolution mapping $g: \mathbb{R}^{N} \times \mathbb{R}^{n} \longrightarrow$ $\mathbb{R}^{N}, n, N \in \mathbb{N},$ and a reservoir readout (output) mapping $f: \mathbb{R}^{N} \rightarrow \mathbb{R}$ that under certain hypotheses transform (or filter) an infinite discrete-time input $\bu=\left(\ldots, \bu_{-1}, \bu_{0}, \bu_{1}, \ldots\right) \in\left(\mathbb{R}^{n}\right)^{\mathbb{Z}}$ into an output signal $\mathbf{y} \in \mathbb{R}^{\mathbb{Z}}$. An expression using the state-space transformation is given by:

\begin{equation}
\label{eq:rcFramework}
\begin{aligned}
\btheta_{t}=g\left(\btheta_{t-1}, \bu_{t}\right) \\
y_{t}=f\left(\btheta_{t}\right)
\end{aligned}
\end{equation}

Although very similar to the RNN family, the evolution function $g$ is not trained using gradient-based methods. Only the readout function $f$ will be trained. Despite this strong architectural simplification, the recurrent part of the model (the reservoir) provides a rich pool of dynamic features by initializing a large reservoir dimension, which are suitable for solving a large variety of tasks, and RC models have already demonstrated excellent performance in many fields, including time series forecasting (see \citet{Bianchi2015}, \citet{Bianchi2015a}, \citet{Deihimi2012}, \citet{Li2012}, \citet{Shi2007}), \citet{li2025llm} process modeling \citep{Rodan2017} and speech analysis \citep{Trentin2015}. In machine learning, due to the discrete time system, RC techniques were originally introduced under the name Echo State Network (ESN) [15]. Studies have shown the superiority of the network in modeling complex, yet lower dimensional nonlinear dynamical system. \citet{Pathak2018a} has demonstrated that ESN is capable of accurately forecasting the chaotic Kuramoto-Sivashinsky dynamical system within 8 Lyapunov times. However, although chaotic, the system is yet deterministic, while the synthetic data generated are essentially ``noise free''. Thus, ESN's performance on stochastic dynamical systems, especially those with bigger and more unstable noise, is still of doubt and less professionally studied. Not surprisingly, in reality, this deterministic model does show a non-robust performance to noise \citep{caonima}. An intuitive explanation would be that the gradient based learning mechanism of ESN treats all output samples $\bm{y}_t$ with equivalent weight. However, the time-varying irreducible noise $\epsilon_t^{ir}$ can sometimes become big enough to deviate and dominate the induced gradient, thus leading parameter learning to a sub-optimal direction. Hence, we are interested in seeking a solution that remedyes RC's susceptibility to noise.

On the other side, the traditional state space models like Kalman filters \citep{Kalman1960} , dynamic linear models (DLMs) \citep{Hodges1991} and their numerous extensions (\citep{Ljung1979},\citep{Wan2000a}) have been utilized for probabilistic inference in time series analysis. This general class of state-space models has been very successful for modeling linear or non-linear stochastic systems. Besides, for a generalized Kalman filter, its state-space representation can be expressed as

\begin{equation}
\label{eq:kalman}
\begin{aligned}
    \btheta_{t}=g\left(\btheta_{t-1}, \bu_{t}, W_t\right) \\
    y_{t}=f\left(\btheta_{t}, \bm{x}_{t}, V_t\right)
\end{aligned}
\end{equation}

Where $W_t, v_t$ are respectively evolution uncertainty random variable and output observation uncertainty random variable. 

%######### Unified

Comparing Equation \ref{eq:rcFramework} and Equation \ref{eq:kalman}, the similarity is significant. This aroused the interest to unify both frameworks to seek for an robust integration that shares good properties of its two parental frameworks; while preserving RC's richness of dynamics and its control-able weight initialization to modulate system's of chaotiveness \citep{Jaeger2010} through the reservoir spectral radius, the integration is robust to system noise and also offers superior uncertainty quantification. Furthermore, due to the proximity of RC and deep sequential models (such as RNN family), such integration can be potentially generalized to more sophisticated deep learning modules such as LSTM \citep{Doerr2018}, sequential architectures \citet{luo2025sequda}, and other gated network structures. Thus, it is of great interest to investigate the feasibility and technical details of such integration. Furthermore, to meet the need of study the integrated model's behavior on noisy time-series, we designate option pricing as the setting to establish a comprehensive study, due to the innate noisiness of option datasets.

%######### Contribution

Our work's contributions are two fold. 

\begin{enumerate}
    \item We present the unscented reservoir smoother (URS), a methodology that offers both uncertainty quantification and rich dynamics in modeling highly noisy data. A comprehensive study on URS' behaviour on synthetic data and forecasting performance testing on real world option data are included, showing that URS functions more superior than its competitors.
    \item We take the gist of URS as building blocks to explore possible extensions of the model. In the end, we are able to demonstrate the potential of fully integrating sequential deep learning model framework and the state-space model framework.
\end{enumerate} 

The remainder of the paper is structured as follows. Section \ref{sec:Notation} presents commons used notations in the rest of the paper. Section \ref{sec:Case_Background} introduces the case background, including necessary technicality in the discipline and its connection to the modeling process. Section \ref{sec:Model_Formulation} formulates the integrated model step-by-step, whereas Section \ref{sec:Param_Inference} provides discussion of inference methodologies on the formulated model. The rest are experiments. In Section \ref{sec:Simulation} and Section \ref{sec:Experiment} we present model evaluation on synthetic data and real-world data, respectively. Lastly, in Section \ref{sec:Discussion} we present possible extensions to our study and its generalizations.

\section{Notation} \label{sec:Notation}

\begin{table}[H]
\begin{tabularx}{\textwidth}{p{0.15\textwidth}X}
\toprule
\textbf{Notation}    & \textbf{Definition} \\
\midrule 
Probability \\
$\cN(\mu, \sigma^2)$    & Normal distribution with mean $\mu$ and variance $\sigma^2$ \\
$\rN(\bm{\mu, \Sigma})$ & Multivariate normal distribution with mean $\bm{\mu}$ and covariance $\bm{\Sigma}$ \\
$\bP, \bE, \bV$         & Probability, expectation, and variance \\

Function \\
$\bG$                   & Reservoir state evolution matrix\\
$\bG_{in}$              & Reservoir input matrix\\
$\bias$                 & Reservoir mean-shift bias\\
$\tau$                  & Reservoir activation function\\
$g$                     & Reservoir state evolution function\\
$f$                     & Reservoir output function; Black-Scholes formula; See Equation \ref{eq:Black_Scholes}\\
$\bm{\Xi}(X)$           & Unscented transformation of $g$ on random variable $X$\\
$\bm{\Psi}(X)$          & Unscented transformation of $f$ on random variable $X$\\

Data \\
$y_{t}^{(i)}$                & The $i^{th}$ Observable Output at timestep $t$; Option Price\\
$\cD_t$                 & $\cD_t := \{y_1, y_2, \dots, y_t\}$ observable output up to $t$\\
$\cD_T$                 & $\cD_T := \{y_1, y_2, \dots, y_T\}$ where $T$ is the last observable timestep\\

Distribution \\
$\bm{m}_t, \bm{C}_t$              & Posterior mean and variance of filtered latent state: $\bP(\btheta_t | \cD_t)$\\
$\bm{m}_t^*, \bm{C}_t^*$          & Smoothed mean and variance latent state: $\bP(\btheta_t | \cD_T)$\\

\bottomrule
\end{tabularx}
\vspace{10pt}
\caption{\small Notation of frequently used parameters}
\label{table:notation}
\end{table}

A list of frequently appearing notations including their corresponding definitions are included in Table \ref{table:notation}. Other notations and definitions will be introduced in relevant paragraphs.

\section{Case Background} \label{sec:Case_Background}
Our objective in this study is to integrate the versatility of reservoir computing and the probabilistic state-space framework to extract latent temporal feature from highly noisy non-linear dynamical systems. We will also run model on real world dataset to test model robustness against high noise. Option data is an ideal candidate, due to its extreme noisiness and the fact that its observable data are calculated from a certain latent volatility process.

\subsection{Option} \label{sec:Option}

Options are derivatives that links to a particular asset. An option is essentially a contract made jointly by the seller and the buyer of the option, specifying an exercise date (Maturity Date), on which the buyer of the option has the right to either buy (call) or sell (put) the linked asset at a pre-specified price (Strike Price). There are 2 different option styles, American and European. An American style option gives the right to the buy side of the option to exercise the option anytime from the purchase date of the option to its maturity date. A European style option, however, limits the buy side of the option to exercise the option only on maturity date. Due to the fact that an option is a derivative to an asset, the value of the option depends on various aspects of the asset, including its risk free rate $r$, spot price $p_t$ and volatility $\sigma^2$. Besides, option prices are also affected by the pre-determined option parameters including strike price $K$ and maturity date $T$. Among these all necessary parameters, all but asset volatility are observable. Volatility, however, needs to be inferred from data using distinctive methodologies, therefore leading to potential huge discrepancies on pricing results on the commensurate option. Such discrepancies on option prices are furthermore recorded into market history data, introducing considerable noise that harshly interferes future statistical inference on instantaneous volatility. We're therefore interested in formulating a model robust and effective in extracting and forecasting volatility trajectory patterns from the highly noisy real-world option market data.

\subsection{Option Pricing: Black-Scholes Model} \label{sec:BS}

\citet*{Black1973} proposed the Black-Scholes Model which gives a theoretical estimate of European-style option price from the parameters stated in the Section \ref{sec:Option}. The model proposes a stochastic partial differential equation, commonly known as the Black-Scholes equation, which describes the instantaneous option price over time:
\begin{align*}
    \frac{\partial y}{\partial t}+\frac{1}{2} \sigma^{2} p^{2} \frac{\partial^{2} y}{\partial p^{2}}+r p \frac{\partial y}{\partial p}-r y &= 0
\end{align*}
Where $y$ is the asset's spot price; $\sigma^2$ is the asset volatility; and $r$ is the risk free interest rate. Black-Scholes formula, on the other hand, is a closed-form evaluation of the partial differential equation above, providing a direct mapping from 5 parameters to the price of call option. The Black-Scholes formula is listed below

\begin{equation} \label{eq:Black_Scholes}
\begin{split}
    y = f(p_t, r, \sigma^2, T, K) &=p_t \bm{\Phi}\left(d_{+}\right) - \bm{\Phi}\left(d_{-}\right) K e^{-r T}  \\
    d_{+} &=\frac{1}{\sigma \sqrt{T}}\left[\ln \left(\frac{p_{t}}{K}\right)+\left(r+\frac{\sigma^{2}}{2}\right)T\right] \\
    d_{-} &=\frac{1}{\sigma \sqrt{T}}\left[\ln \left(\frac{p_{t}}{K}\right)+\left(r-\frac{\sigma^{2}}{2}\right)T\right] 
\end{split}
\end{equation}

Where
\begin{itemize}[itemsep=0pt]
    \item $p_t$ is the spot price of the underlying asset at time $t$.
    \item $r$ is the risk free interest rate.
    \item $\sigma^2$ is quadratic of the volatility of the underlying asset.
    \item $K$ is the strike price of the call option. 
    \item $T$ is the maturity time since $t$ of the option.
    \item $\bm{\Phi}(\cdot)$ is cumulative distribution function of standard normal distribution.
\end{itemize}

Black-Scholes formula offers pricing only for call option, whereas the corresponding put option price is calculated using the Put-call parity \citep{Stoll1969} equation to ensure absence of potential arbitrage in the market. The Put-call parity equation is out of the scope of our study's interest and focus.

Notice that Black-Scholes formula only incorporate constant $r$ risk free rate and constant $\sigma^2$ volatility, the Black-Scholes is constructed upon strong assumptions. First, the model assumes that the asset holds a constant risk-free return, which is the risk-free rate $r$; Second, the model assumes asset price to follow a geometric Brownian motion. Therefore, the log return of the asset is a random walk process with constant variance. To this reason, standard deviation of the asset's log price process, commonly known as volatility $\sigma$, is also assumed to be constant.

In the real world situation, Black-Scholes formula including its related extensions are still widely considered for pricing an option. However, as $\sigma^2$ is not latent and unobserved, an estimated or implied volatility $\hat{\sigma}^2$ must be calculated from past asset price data or past option data. However, the inferred $\hat{\sigma}^2$ and actual risk free rate $r$ need not be constant over time. Due to this reason, pricing models needs to constantly reprice the option by utilizing updated $\hat{\sigma}^2$ and $r$. But even so, to price using Black-Scholes is to acquiesce a constant volatility and risk free rate of that asset since the pricing timestep. Due to this insufficiency of Black-Scholes and other reasons such as volatility smile of an asset, real world market data need not strictly obey Black-Scholes formula, as the pricing data might be resulted from calibrated volatility or option price. Such nuance can harshly impact how we conduct real world data model comparison. Details will be further discussed in Section \ref{sec:Test_Method}.

To break the strong assumptions made by Black-Scholes model, rough stochastic models such as the Heston model \citep{Heston1993}, see Equation \ref{eq:Heston}, which breaks the constant volatility process assumption by introducing a rough CIR volatility process to replace that \citep{Cox1985}, are further proposed. Such models have the ability to achieve better pricing results if inference of parameters used in stochastic partial differential equations are conducted properly. However, we decided to stick with Black-Scholes formula the pricing function, though implementing more complex stochastic volatility pricing functions may help improve the result of our model. This is for the purpose of simplicity and concentrate on comparing our model's performance on filtering, inferring, and forecasting latent volatility process to that of others.

\begin{equation} \label{eq:Heston}
    \begin{aligned}
        d p_{t} &= \mu p_{t} d t+\sqrt{v_{t}} p_{t} d W_{t}^{(p)}\\
        d v_{t} &= \kappa\left(\theta-v_{t}\right) d t+\xi \sqrt{v_{t}} d W_{t}^{(v)}\\
        d W_{t}^{(p)}W_{t}^{(v)} &= \rho dt
    \end{aligned} 
\end{equation}

\section{Model Formulation} \label{sec:Model_Formulation}

We first want to formulate a complete filtering of nonlinear dynamical system problem in the context of our case study. From market data, we observe a series of option price vectors $\bm{y}_t$. Each vector $\bm{y}_t$ is comprised of the top $I$ most traded options $y_{t}^{(i)}, i \in\{1,2, \dots I\}$ quotes on day $t$. The value of $y_{t}^{(i)}, \forall i \in {1,2, \dots, I}$ is determined by Black-Scholes formula with ulterior parameters $r_t, v_t, p_t, K_{t}^{(i)}, T_{t}^{(i)}$ with stated meaning in the previous Section \ref{sec:BS}. 

\begin{equation} \label{eq:ModelBlackScholes}
\begin{aligned}
    y_{t}^{(i)} = f(p_t, r_t, \sigma_t^2, T_{t}^{(i)}, K_{t}^{(i)}) &=p_t \bm{\Phi}\left(d_{+}\right) - \bm{\Phi}\left(d_{-}\right) K_{t}^{(i)} e^{-r_t T_{t}^{(i)}}  \\
    d_{+} &=\frac{1}{\sigma_t\sqrt{T_{t}^{(i)}}}\left[\ln  \left(\frac{p_{t}}{K_{t}^{(i)}}\right)+\left(r_t+\frac{\sigma_t^2}{2}\right)T_{t}^{(i)}\right] \\
    d_{-} &=\frac{1}{\sigma_t\sqrt{ T_{t}^{(i)}}}\left[\ln  \left(\frac{p_{t}}{K_{t}^{(i)}}\right)+\left(r_t-\frac{\sigma_t^2}{2}\right)T_{t}^{(i)}\right] 
\end{aligned}
\end{equation}

Among the 5 parameters, volatility $\sigma_t$ is then computed from a latent reservoir state vector $\bm{\theta}_t$ via function $h$. Finally, the $\bm{\theta}_t$ is updated from the state of the previous timestep $\bm{\theta}_{t-1}$ together with a input control vector $\bm{u}_{t}$ via a predetermined function $g$.

Above summarizes the model structure, which is essentially a state-space model filtering of dynamical system problem. The formal model is defined as below, where the state evolution function $g$ is modeled by an echo state neural network, observation output function $f$ is modeled by the Black-Scholes formula, and the probabilistic setting fits into the unscented Kalman filter framework.

\begin{align*}
    \bu_t &= [\frac{p_{t-m} - p_{t-m-1}}{p_{t-m-1}}, \frac{p_{t-m+1} - p_{t-m}}{p_{t-m}}, \cdots, \frac{p_{t} - p_{t-1}}{p_{t-1}}] \\
    \btheta_t &= g(\btheta_{t-1}, \bu_{t}, W_t)\\
    \sigma_t &= h(\btheta_t)\\
    y_{t}^{(i)} &= f_{t}^{(i)}(p_t, r_t, \sigma_t^2, T_{t}^{(i)}, K_{t}^{(i)}) + \nu_t  \\
    g(\btheta_{t-1}, \bu_{t}, W_t) &= \tau(\bG \btheta_{t-1} + \bG_{in} \bu_t^2 + \bias) + W_t \\
    h(\btheta_{t}) &= \frac{1}{p}\sum_{i=1}^p \theta_{ti} \\ 
    f_{t}^{(i)}(p_t, r_t, \sigma_t^2, T_{t}^{(i)}, K_{t}^{(i)}) &= \text{Equation \ref{eq:ModelBlackScholes}}  \\
    W_t &\sim \rN(\bm{0}, \bm{W}) \\
    \nu_t &\sim \cN(0, v)
\end{align*}

We will proceed into the discussion of echo state neural network and \textit{\textbf{unscented Kalman filter (UKF)}} respectively to formulate each part of our formal model, backed with relevant definitions and derivations in the rest of this section. Methodologies of parameter inference and training algorithm will be further discussed and displayed in Section \ref{sec:Param_Inference}.

\subsection{Echo State Neural Network} \label{sec:ESN}

\textit{\textbf{Echo state neural network (ESN)}} (\citet{Jaeger2002}, \citet{Jaeger2007}) is a reservoir computing approach (\citet{Jaeger2004}, \citet{Maass2000}, \citet{Maass2002}, \citet{Crook2007}, \citet{Verstraeten2007}, \citet{Lukosevicius2009}) capable of generating non-linear system of dynamics. The most simple ESN implements the following transition and output functions:

\begin{equation} \label{eq:ESN}
\begin{aligned}
    \btheta_t &= g(\btheta_{t-1}, \bm{u}_t) \\
    &= \tau(\bG \btheta_{t-1}  + \bG_{in} \bm{u}_{t}^2) \\
    \tau(x) &= \frac{1}{1 + e^{-x}}
\end{aligned}
\end{equation}

$\bG$ is reservoir state evolution matrix, and $\bG_{in}$ is the reservoir control input weight. An intuitive illustration of the model is shown in Figure 1. Notice that $\sigma$ can be any sigmoidal function; Traditionally, function $\sigma$ is default set to tanh function \citep{Jaeger2010}. However, we're setting the activation to be logistic function due to the fact that volatility lies in the range of $(0,1)$. We will first discuss weight initialization design to ensure the \textit{\textbf{echo state property} (ESP)} (see \citet{Jaeger2010}, \citet{Manjunath2013}) in Section \ref{sec:ESP}, and will then further discuss design of function $h$, activation function $\tau$, weight initialization of $\bG_{in}$, and other components in the echo state neural network in Section \ref{sec:Modification_to_Model}.

\subsubsection{Echo State Property} \label{sec:ESP}

An ESN relies on a reservoir evolution function $g$ that must satisfies the echo state property (ESP) to vanish long-term state initialization memory. To establish echo state property, the input control sequence $\bm{u}_{1:t}$ and reservoir state $\btheta_t$ should lie in compact domains :

\begin{figure}[H]
    \begin{center}
        \vspace{0.25in}
        \includegraphics[width=0.6\linewidth]{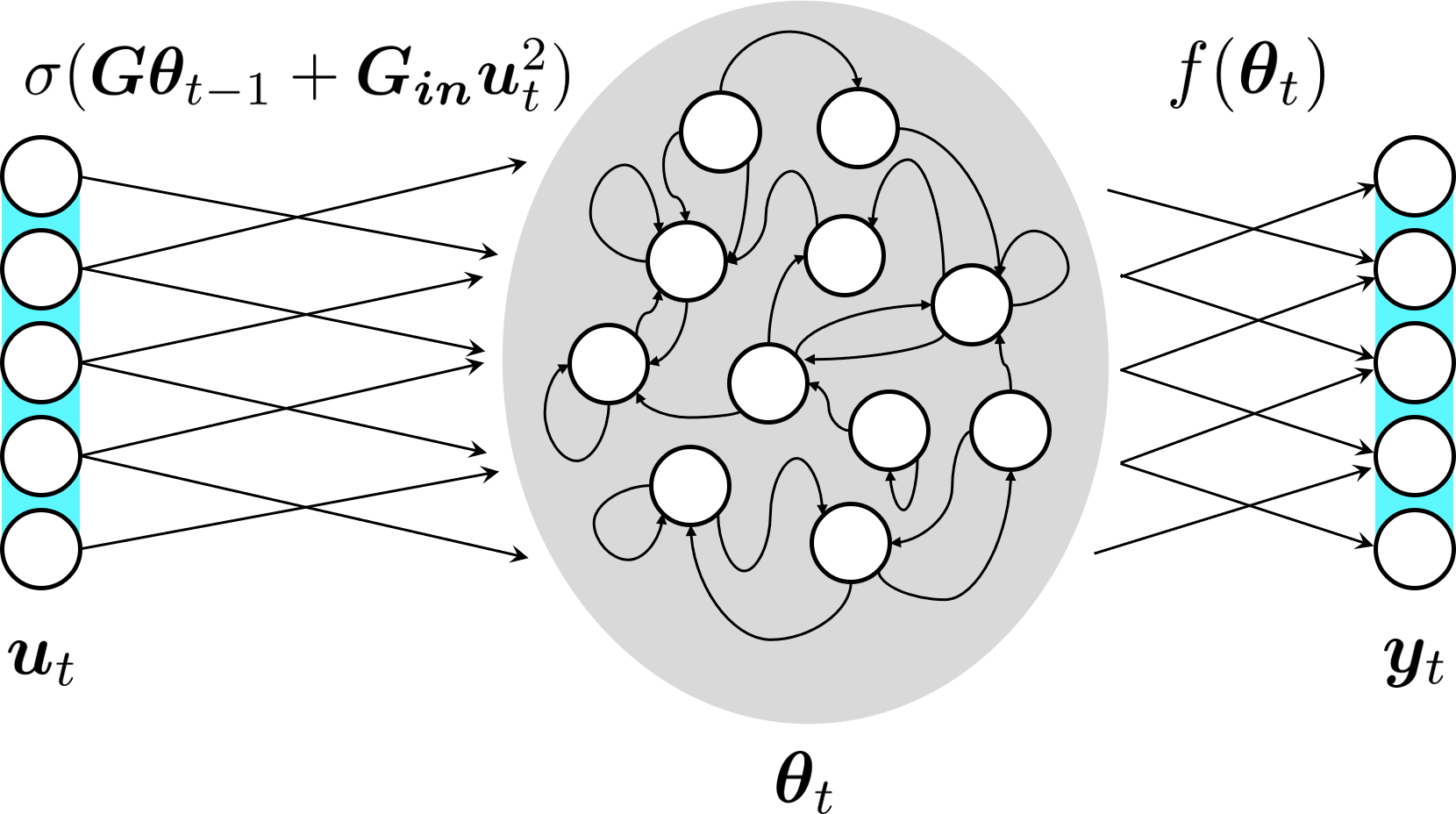}
        \vspace{10pt}
    \end{center}
    \figurecaption{Figure \ref{fig:ESN_Structure}}{A typical echo state neural network structure. In the most fundamental ESN structure, direct output function from input $\bu_t$ and feedback mapping from $\bm{y}_t$ are not included.}
    \vspace{0.25in}
    \captionlistentry{}
    \label{fig:ESN_Structure}
\end{figure}

\begin{defi} 
    Given a nonlinear dynamical systems equipped with transition function $\bm{\theta}_t = g(\bm{\theta}_{t-1}, \bm{u}_t)$, where $\bm{u}_t \in \mathcal{U}, \forall t \in \mathbb{N}$ is the control input sequence that lies in a compact domain $\mathcal{U}$. If all admissible hidden state $\bm{\theta}_t \in \mathcal{H}, \forall t \in \mathbb{N}$, where $\mathcal{H}$ is compact, the system is said to satisfy \textbf{standard compactness conditions}.
\end{defi}

An alternative way of phrasing the definition is that, whenever the system starts, for instance at step $t'$, $\bm{\theta}_t \in \mathcal{H}$, and $\forall \bm{u}_{t+1} \in \mathcal{U}$, the projected state space estimate $\bm{\theta}_{t+1} \in \mathcal{U}$. Namely, the system state is closed under evolution function $g$.

\begin{defi}
    Given a Recurrent Neural Network equipped with transition function $g$ that satisfies the standard compactness condition. If the state $\bm{\theta}_{t}$ is computed solely on $\bm{\theta}_{t-1}$ and $\bm{u}_{t}$ (i.e. $g$ takes in only $\bm{\theta}_{t-1}$, $\bm{u}_{t}$ as the 2 parameters). Furthermore, if $\bm{\theta}_{t}$ is uniquely determined by any left-infinite input sequence $\bm{u}_{[-\infty: t]}$, and then network satisfies \textbf{echo state property}.
\end{defi}

To rephrase definition 2, we consider a network with with transition $g$ that only takes in $\btheta_{t-1}, \bu_{t}$ as input. The echo state property is satisfied only if and only if given any fixed left-infinite control input sequence $\bu_{[-\infty : T]} \in \mathcal{U}$, given any 2 left-infinite state sequence $\btheta_{[-\infty : T]},\  \btheta'_{[-\infty : T]}$, where 

\[
\btheta_{t} = g(\btheta_{t-1}, \bu_{t}) \quad \btheta'_{t} = g(\btheta'_{t-1}, \bu_{t}) \quad  \forall t \leq T \in \mathbb{Z}
\]

We can deduce $\btheta_t = \btheta'_t$. Namely, echo state property is satisfied when an ESN's long-term state estimate is determined only by the  control input sequence. Discrepancy in initialization of state $\btheta_t' \neq \btheta_t$ will eventually vanish, leading to $\lim_{t\to \infty}(\btheta_t' - \btheta_t) = 0$. Thus, dynamics generated by echo state neural network satisfying ESP is robust to arbitrary state initialization given a long enough control sequence $\bm{u}_{0:t}$, as the system is state forgetting and will eventually wash out errors caused by initialization and contract to a limiting dynamic. Such property aligns with our case background, where the initial different market estimates of volatility will be guaranteed to converge and agree in a long term.

Next, we're interested in constructing an ESN with ensured ESP. To have ESP, we need to ensure that $\forall \btheta_{[0:t]}, \btheta'_{[0:t]}$, the $l_1$ norm $\| \btheta_t - \btheta'_t \|_{L_1} \to 0$ as $t \to +\infty$. Though an equivalent statement as ESP is hard to find, we can place stronger condition to ensure ESP. A generic way to achieve ESP has been proposed by Jaeger in 2010 by limiting the reservoir state evolution matrix to have spectral radius $\rho < 1$:

\begin{prop} \citep{Jaeger2010}
    Consider an echo state neural network with $\rho(\bG) < 1$. Suppose the network's sigmoidal function is implemented as logistic function
    \[
    \tau(x) = \frac{1}{1 + e^{-x}}
    \]
    Then, $\forall t \in \mathbb{N}$, $\forall \bm{u}_{t_0:+\infty} $, $\forall \btheta_t \neq \btheta'_t \in \mathcal{H}$ compact, we have 
    \[
    \|\btheta_{t+1} - \btheta'_{t+1} \|_{L_1} = \| g(\btheta_t, \bu_t) - g(\btheta'_t, \bu_t) \|_{L_1} < \|\btheta_{t} - \btheta'_{t} \|_{L_1} \] 
\end{prop}

\begin{proof}
    It is easy to show that $\forall a, b, u \in \mathbb{R}$, $|g(a, u)- g(b, u)| < |a-b|$, as $\limsup{g'} = \frac{1}{2} < 1$. $g$ is a contraction on $\mathbb{R}$. Then we can show that for $\btheta \in \mathbb{R}^p$:
\begin{align*}
    \|\btheta_{t+1} - \btheta'_{t+1} \|_{L_1} &= \| \tau(\bG\btheta_{t}+\bG_{in} \bm{u}_t^2) - \tau( \bG\btheta'_{t} + \bG_{in} \bm{u}_t^2) \|_{L_1} \\
    &< \| (\bG\btheta_{t}+\bG_{in} \bm{u}_t^2) - ( \bG\btheta'_{t} + \bG_{in} \bm{u}_t^2) \|_{L_1} \\
    &= \| \bG(\btheta_{t} - \btheta'_{t})\|_{L_1} \\
    &\leq \rho(\bG)\| (\btheta_{t} - \btheta'_{t})\|_{L_1} 
\end{align*}
Thus, when $\rho(\bG) < 1$, $\|\btheta_{t}'-\btheta_{t}'\|$ shrinks in the power of $\rho(\bG)$. Thus, $\|\btheta_{t}'-\btheta_{t}'\| \to 0$
\end{proof}

From the proposition above, it suffices to initialize and adjust the spectral radius to be smaller than 1 to ensure echo state property of the reservoir. 

\subsubsection{Modification to the model} \label{sec:Modification_to_Model}

In our case, we'd specify instantaneous volatility to be calculated from $h(\btheta_t)$. Due to identifiability issue and interpretability purposes, we configure $h$ to take the mean of each entry of $\btheta_t$ to calculate $\sigma_t$ as the standard deviation representation of volatility. Averaging assigns the equal weight to each dimension of the reservoir dynamics. Therefore, while resolving identifiability issue introduced by approximating the composite of 2 uncertain functions $h \circ g$, it also help identify the amount of contribution made by each reservoir dynamic $\theta_{ti}$ by directly comparing the magnitude of $\theta_{ti}$ at each time step. 

\begin{align*}
    h(\btheta_t) := \frac{1}{p}\sum_{i=1}^p \theta_{ti} = \sigma_t
\end{align*}

As mentioned previously, all sigmoidal functions can be used for reservoir activation. However, a nonlinear activation should have function range that matches the targeting value range. With $h$ already specified, we understand that $\forall i \in {1,2,\cdots, p}, \theta_{ti} > 0$ is a requirement as the volatility $\sigma_t$ is a positive value. Furthermore, annualized volatility wouldn't be larger than 1 ($100\%$). Therefore, all the potential value lies in the admissible state space given the logistic activation function.

Besides, we're interested in ensuring the stability along $\btheta_t$ process when mapping $\btheta_t \in (0,1)$ to $\btheta_{t+1}$ using function $g$. Unfortunately, the dynamical system requires more than $\bG, \bG_{in}$ to ensure stability. For instance: instantaneous volatility ranges around $0.1$ to $0.2$. Mapping $\btheta_t \in  (0.1, 0.2)$ to $\btheta_{t+1} \in (0.1, 0.2)$ is non-feasible without a mean-shift for the function, as the $g((0.1, 0.2)) \cap (0.1, 0.2) = \emptyset$. Thus, a negative mean-shift bias should be applied within the logistic function to not harm ESP, while shifting the logistic function to along $x$-axis to the negative side, therefore ensures a ``domain-range match'' of the reservoir state space. 

Finally, a reasonable scaling factor of $\bG_{in}$ should be explored in order to efficiently conduct further parameter inference. For example, in our case, the daily return of an SP500 index is usually within the level of $10^{-2}$. Furthermore, due to the positive relation between quadratic of return and volatility, we use quadratic of $\bm{u}_t^2$ instead of $\bm{u}_t$ as input (as already shown above in Equation \ref{eq:ESN} Taking quadratic leads to a diminutive variation within the input control sequence, leaving its control impact on the model system very weak. Therefore, either a scaling on $\bm{u}_t^2$ or $\bG_{in}$ or both need to be applied. Although an arbitrary scaling empirically won't harm the system's echo state property, but a reasonable scaling will significantly boost efficiency for the parameter inference step, which includes the gradient based generalized EM algorithm.

Therefore, the finalized state evolution function is as follows. Detailed weight initialization procedure is described in Algorithm \ref{algo:ESN_Init} \citep{Lukosevicius2012}.

\begin{equation} \label{eq:FinalEvolution}
    \begin{aligned}
        \btheta_{t} &= g(\btheta_{t-1}, \bm{u}_t) \\
        &= \tau(\bG\btheta_{t-1} + \bG_{in} \bm{u}_{t}^2 + \bias \\
        \tau(x) &= \frac{1}{1+ e^{-x}}
    \end{aligned}
\end{equation}

\begin{algorithm}[h]
\caption{Weight initialization for ESN reservoir}
\label{algo:ESN_Init}
\SetAlgoLined
\KwResult{A short algorithm of initializing reservoir weights with insured echo state property. Through empirical experiments, parameters are recommended to be set to $\eta_1 = 0.97, \eta_2 = 0.85, \bm{\mu} = -2, \epsilon=1 $ . }
 \textbf{Initialize}: weight of $\bG$, $\bG_{in}$ randomly; \\
 \textbf{Calculate}: the spectral radius $\rho$ of $\bG$; \\
 \textbf{Set:} $\bG := \eta_1 \bG / \rho  $, where $\eta_1 < 1$ is the evolution shrinkage factor; \\ 
 \textbf{Set}: $\bG_{in} := \eta_2 \bG_{in} $, where $\eta_2 < 1$ is the input shrinkage factor; \\
 \textbf{Initialize}: $\bias \sim \rN(\bm{\mu}, \epsilon \bm{I})$; \\
 \textbf{Return}: $\bG, \bG_{in}, \bias$
\end{algorithm}

\subsection{Unscented Kalman Filter} \label{sec:UKF}

ESN is a deterministic modeling process, apart from potentially training with noise-added feedback from output \citep{Jaeger2005}. To provide further uncertainty measurement for latent reservoir states $\btheta_t$, we seek to cast ESN framework into the Bayesian Kalman filter framework. This will offer filtered and smoothed uncertainty measurements and also increase learning robustness to in-sample error to improve robustness to noise \citep{Agamennoni2011}. Equation \ref{eq:KalmanFramework} proposes a generalized state-space model framework with not necessarily linear evolution function, where $\bm{u}_t$ is the input control sequence, and $\bm{o}_t$ are exogenous configuration parameters for function $f$, providing extra variability in the output function.

\begin{equation} \label{eq:KalmanFramework}
\begin{aligned}
    \btheta_t &= g(\btheta_{t-1}, \bm{u}_{t}, W_t)\\
    y_{t} &= f_{\bm{o}_t}(\btheta_t, \bm{o}_t, \nu_t )  \\
    W_t &\sim \rN(\bm{0}, \bm{W}) \\
    \nu_t &\sim \cN(0, v)
\end{aligned}
\end{equation}

By close observation, we found that in general, the reservoir computing framework can be cast into the Kalman filtering framework. Due to this similarity, uncertainty quantification and Bayesian update mechanism could be incorporated into the original deterministic modeling and predicting procedure. This will benefit the deterministic ESN model in 3 ways:

\begin{enumerate}
    \item Robustness to noisy data: the ESN approach itself tune model parameters by minimizing equally weighted error induced by all training samples. However, due to the fact that output observation $\bm{y}_t$ are generated from noisy and several biased reservoir states estimate $\hat{\btheta}_t$, there can be drastic variation of irreducible error $\epsilon^{(ir)}_t$ within all the output $\bm{y}_t$. Suppose that the irreducible error $\epsilon^{(ir)}_t$ dramatically increase in some $\bm{y}_t$ for some $t$, information drawn from $\bm{y}_t$ should acquire less weight in influencing the model parameter, as higher $\epsilon^{(ir)}_t$ indicates higher uncertainty of information that could be drawn from this timestep. However, deterministic MSE assigns identical weight to all time steps. This can impair learning. Moreover, bigger $\epsilon^{(ir)}_t$ usually leads to more biased $\bm{y}_t$. Therefore equal weighting of $\bm{y}_t$ will introduce more learning error in adjusting the model parameter and results in more biased parameter estimate. Thus, modeling and learning should incorporate mechanism in self-adjust learning weight. Bayesian prior and posterior update mechanism from Kalman filter framework is a solution in achieving the desired property, as the update incorporate data precision (inverse of sample variance) from $\bm{y}_t$ to measure ``credibility'', therefore adjust the learning weight correspond to it. Informally and intuitively speaking, Kalman filter framework enables the ESN to learn from some ``credible'' $\bm{y}_t$ output, while ``doubt'' the those $\bm{y}_t$ output that are more uncertain. This is further discussed in Section \ref{sec:FF} .
    \item Potential for online learning: unlike deterministic ESN that requires fitting output linear weight every time the next timestep measurement is collected, Kalman filter can adjust future prediction simply via updating the reservoir state estimate through prior posterior update. Furthermore, by applying joint Kalman filter, $\bG$, $\bG_{in}$ can simultaneously updated with $\btheta_t$. This enables this model to learning online. Details will be described in Section \ref{sec:Online} .
    \item Uncertainty measurement: the Kalman filter framework offers interval estimate of observed states and also predicted state. This shed light on uncertainty of the volatility process $\sigma_t$.
\end{enumerate}

To formally cast the reservoir into the Kalman filter framework, we finalizes the model as follows:

\begin{align*}
    \bu_t &= [\frac{p_{t-m} - p_{t-m-1}}{p_{t-m-1}}, \frac{p_{t-m+1} - p_{t-m}}{p_{t-m}}, \cdots, \frac{p_{t} - p_{t-1}}{p_{t-1}}] \\
    \btheta_t &= g(\btheta_{t-1}, \bu_{t}, W_t)\\
    \sigma_t &= h(\btheta_t)\\
    y_{t}^{(i)} &= f_{t}^{(i)}(p_t, r_t, \sigma_t^2, T_{t}^{(i)}, K_{t}^{(i)}) + \nu_t  \\
    g(\btheta_{t-1}, \bu_{t}, W_t) &= \tau(\bG \btheta_{t-1} + \bG_{in} \bu_t^2 + \bias) + W_t \\
    h(\btheta_{t}) &= \frac{1}{p}\sum_{i=1}^p \theta_{ti} \\ 
    f_{t}^{(i)}(p_t, r_t, \sigma_t^2, T_{t}^{(i)}, K_{t}^{(i)}) &= \text{Equation \ref{eq:ModelBlackScholes}}  \\
    W_t &\sim \rN(\bm{0}, \bm{W}) \\
    \nu_t &\sim \cN(0, v)
\end{align*}

\subsubsection{Unscented Transform} \label{sec:UT}

The mechanics of Kalman filtering (\citet{Kalman1960}, \citet{West1999})  requires the predicted priori state distribution to be Gaussian. In a classical Kalman filter setting, as state transformation is linear, the Gaussian identity of distribution $\btheta_t | \cD_t$ will be preserved under the transformation. However, under nonlinear transformation, such Gaussian identity is not preserved and therefore stops the Kalman filtering procedure. Thus, it is needed to find a Gaussian distribution that well approximate the transformed distribution. An computational effective yet relatively precise method is unscented transform.

The \textit{\textbf{unscented transformation (UT)}} (\citet{Wan2000}, \citet{Wan2000}, \citet{Julier2004a}) approximates a nonlinear-transformed random variable using Gaussian distribution by tracking a set of sigma points along the transformation. Consider a $p$ dimensional multivariate Gaussian Distributed random variable $\btheta_t$ undergoes nonlinear transformation $g$ 

\begin{align*}
    \btheta_t &\sim \rN_p(\bM_t, \bC_t) \\
    \btheta_{t+1} &= g(\btheta_t)
\end{align*}

Then, the optimal set of $ 2p+1 $ sigma points $\{\mathcal{P}_0, \mathcal{P}_1, \cdots, \mathcal{P}_{2p}\}$ required to estimate the targeting distribution $\btheta_{t+1}$ are calculated based on distribution mean and covariance of $\btheta_t$. Among which, the first sigma point is selected to be the mean of $\btheta_t$, and the rest branches around the mean as defined below. Here, the foot note $[i, :]$ denotes the $i^{th}$ row of the matrix. 

\begin{align*}
\mathcal{P}_{0} &= \bM_t \\
\mathcal{P}_{i} &= \bM_t + \left(\sqrt{(p+\lambda) \bC_t}\right)_{[i, :]}  \quad\quad\quad i \in \{1, \ldots, p\} \\
\mathcal{P}_{i} &= \bM_t - \left(\sqrt{(p+\lambda) \bC_t}\right)_{[i-p, :]} \quad\quad i \in \{p+1, \ldots, 2 p\}
\end{align*}

Then, each sigma point is assigned with a pair of weights, one for calculating transformed mean, $\mathcal{W}^{(m)}$, while the other for calculating transformed covariance, $\mathcal{W}^{(c)}$. 

\begin{align*}
\mathcal{W}_{0}^{(m)} &=\lambda /( p +\lambda) \\
\mathcal{W}_{0}^{(c)} &=\lambda /( p +\lambda)+\left(1-\alpha^{2}+\beta\right) \\
\mathcal{W}_{i}^{(m)} &= 1 / \{2( p +\lambda)\} \quad i=1, \ldots, 2 p \\
\mathcal{W}_{i}^{(c)} &= 1 / \{2( p +\lambda)\} \quad i=1, \ldots, 2 p 
\end{align*}

In the expression above, $\lambda$, $\alpha$, $\beta$ are hyper-parameters. $\lambda$ denotes the scaling of sigma points. A typical way of setting these parameters has been proposed by \citet{Wan2000} that $\lambda = \alpha^2(p +\kappa) - p$, $\alpha = 0.001$,  $\kappa = 0$, and $\beta = 2$. We follow their suggested practice for initialization. Then, each sigma points is transformed by function $g$, and the transformed random variable's mean and covariance can be estimated as follow:

\begin{align*}
    \mathcal{Q}_i &= g(\mathcal{P}_i) \\
    \widehat{\bE[\btheta_{t+1}]} &= \sum_{i=0}^{2p} \mathcal{W}_i^{(m)} \mathcal{Q}_i\\
    \widehat{\bV[\btheta_{t+1}]} &= \sum_{i=0}^{2p} \mathcal{W}_i^{(c)}( \mathcal{Q}_i-\widehat{\bE[\btheta_{t+1}]})( \mathcal{Q}_i-\widehat{\bE[\btheta_{t+1}]})^T
\end{align*}

Unscented transform is an efficient algorithm to approximate a transformed random variable by a Gaussian random variable. When applying reservoir computing in an Bayesian framework, the large number of parameter to be inferred makes consistent sampling approach such as particle filter computationally expensive. Thus, unscented transform places a good balance between estimation precision and computation.

\subsubsection{Forward Filtering} \label{sec:FF}

Unscented forward filtering shares similar procedure as that of Kalman filter. Denote $\cD_t$ as all the information carried by $\{y_1, y_2, \cdots, y_t\}$. The model make prediction on next step latent state $\btheta_{t+1} | \cD_t$ based on current posterior distribution of $\btheta_t | \cD_t$ via unscented transform. 

\begin{align*}
    \btheta_t | \cD_t &\sim \rN(\bm{m}_t, \bm{C}_t) \\
    g(\btheta_t | \cD_t ) &\approx \Xi (\btheta_t | \cD_t) \\
    &\sim \rN( \Xi_{\bm{\mu}} [\btheta_t | \cD_t], \Xi_{\bm{\Sigma}} [\btheta_t | \cD_t] ) \\
    \btheta_{t+1} | \cD_t &= \rN(\Xi_{\bm{\mu}} [\btheta_t | \cD_t], \Xi_{\bm{\Sigma}} [\btheta_t | \cD_t]+\bm{W} )\\
    &= \rN( \bm{a}_{t+1}, \bm{R}_{t+1})
\end{align*}

Here, $\Xi$ denotes the unscented transform that approximates $g$, the reservoir state evolution function. Then, output distribution $y_{t+1} | \cD_t$ is again estimated using Unscented transform of $f$.

\begin{align*}
    \btheta_{t+1} | \cD_t &\sim \rN(\bm{a}_{t+1}, \bm{R}_{t+1}) \\
    \sigma_{t+1} | \cD_t &\sim \cN(\frac{\bm{a}_{t+1}}{p}, \frac{\sum_{i,j}\bm{R}_{t+1}}{p^2})\\
    f(\sigma_{t+1} | \cD_t ) &\approx \Psi (\sigma_{t+1} | \cD_t) \\
    &\sim \rN( \Psi_{\bm{\mu}} [\sigma_{t+1} | \cD_t], \Psi_{\bm{\Sigma}} [\sigma_{t+1} | \cD_t] ) 
\end{align*}

Therefore, we can perform Bayesian update by

\begin{align*}
    \bP(\btheta_{t+1} | \cD_{t+1}) &= \frac{ \bP(\btheta_{t+1} | \cD_{t})\prod_{i=1}^k \bP( y_{t+1, i} | \btheta_{t+1} , \cD_t )}{\prod_{i=1}^k \bP(y_{t+1,i} | \cD_{t})}\\
    &\propto \bP(\btheta_{t+1} | \cD_{t})\prod_{i=1}^k \bP( y_{t+1, i} | \btheta_{t+1} , \cD_t ) 
\end{align*}

Where the prior and likelihood are both normal distribution. Notice that during normal-normal update, the inverse of prior variance $\bV(\btheta_{t+1} |\cD_t)^{-1}$ and the observed inverse of data quadratic deviation from mean $(y_{t+1,i} - \bE[y_{t+1, i} | \btheta_{t+1}, \cD_t])^{-2}$ are represented as ``weights'' to calculate weighted average of the prior mean $\bE(\btheta_{t+1} |\cD_t) = \bm{a}_{t+1}$ and observed data mean . Therefore, when irreducible error $\epsilon_{t+1}^{(ir)}$ is high, the later is low, meaning a less weight on data observation and therefore less information drawn from $y_{t+1, i}$. Thus, latent process $\btheta_{t+1}$ remains inert to be deviated from $y_{t+1, i}$. 

Different from of extended Kalman filter, unscented Kalman filter's complex transition function to estimate predicted priori distribution is nonetheless non-closed-form. Due to this reason, the common Kalman filter update equation is not available. Though forward filtering has been shown to be feasible, the absence of tracking cross covariance between $\btheta_t, \btheta_{t+1} | \cD_t$ can cause potential problem for backward smoothing and parameter inference, which will be introduced later in Section \ref{sec:Back_Smoothing} and \ref{sec:Offline}.

\subsubsection{Backward Smoothing} \label{sec:Back_Smoothing}

Forward filtering enables us to excavate posterior distribution of $\btheta$ given up to time $t$. However, we'd also be intrigued by implications from future observations, $\bm{y}_{t+1 : T}$ on previous latent state distribution---that is, we're interested in the distribution of $\btheta_t | \cD_T$, where $T > t$ denotes the last time step of all the observable data. This procedure encourages all state estimates to account for global likelihood of the observed output $\bm{y}_{1:T}$. To achieve this, we walk over the derivation of Unscented RTS smoother and its extension \citep{Sarkka2008}.

Given $\cD_{T} := \{y_1, y_2, y_3, \dots, y_T\}$ the entire dataset, through forward filtering, we've obtained:

\begin{align*}
    \btheta_t | \cD_t &\sim \rN(\bm{m}_t, \bm{C}_t)\\
    \btheta_{t+1} | \cD_t &\sim \rN(\bm{a}_{t+1}, \bm{R}_{t+1})
\end{align*}

Observe that for the last state estimate $\btheta_T$, $\btheta_T | \cD_T$ is obtainable solely from forward filtering as $T$ is already the last time step, we can start from $\btheta_T | \cD_T$, finding relationship between $\btheta_t | \cD_T$ and $\btheta_{t-1} | \cD_T$, and solving pairwise joint distribution of $(\btheta_{t} | \cD_T, \btheta_{t-1} | \cD_T)^T$ recursively $\forall t \in \{1, 2, \cdots, T\}$. By induction, assume we already know that 

\[
\btheta_{t} | \cD_T \sim \rN(\bm{m}^*_t, \bm{C}^*_t)
\]

We are interested in calculating the joint distribution of $(\btheta_t, \btheta_{t+1})$. Unfortunately, due to that closed form transition function from $\btheta_t$ to  $\btheta_{t+1}$ is not available, we cannot track the cross-covariance between the 2 consecutive random variables, comparing to extended Kalman smoothing \citep{Psiaki2005} where Taylor expanding the non-linear transformation can provide a approximated yet closed cross-covariance. However, unscented transform can overcome this difficulty. We could track cross-covariance through concatenating $\btheta_{t}$ to itself in a $2p$ dimensional vector, and perform $g'$ as shown in Equation \ref{eq:trackCrossCov} to represent $g$ by a partial transformation on this random variable. This will not only track the cross-covariance, but also keeps the fidelity of the cross-covariance estimation:

\[
\left.
\begin{bmatrix}
\btheta_t \\
\btheta_t
\end{bmatrix}
\right| \cD_t
\sim \rN \left(
\begin{bmatrix}
\bm{m}_t \\
\bm{m}_t
\end{bmatrix},
\begin{bmatrix}
\bm{C}_t & \bm{C}_t \\
\bm{C}_t^T & \bm{C}_t
\end{bmatrix}
\right)
\]

\begin{equation} \label{eq:trackCrossCov}
    \begin{aligned}
        \left. \begin{bmatrix}
        \btheta_t \\
        \btheta_{t+1}
        \end{bmatrix} \right | \cD_t &=  \left. \begin{bmatrix}
        \btheta_t \\
        g(\btheta_{t}) + W_t
        \end{bmatrix} \right | \cD_t 
        \\
        = g'\left( \left. \begin{bmatrix}
        \btheta_t \\
        \btheta_{t}
        \end{bmatrix} \right | \cD_t \right) 
        +
        \begin{bmatrix}
        0 \\
        W_t
        \end{bmatrix}
        &\sim \rN \left(
        \begin{bmatrix}
        \bm{m}_t \\
        \Xi_{\bm{\mu}}(\btheta_t)
        \end{bmatrix},
        \begin{bmatrix}
        \bm{C}_t & \bm{\Sigma}_{t, t+1} \\
        \bm{\Sigma}_{t+1, t} & \Xi_{\bm{\Sigma}}(\btheta_t) + \bm{W}
        \end{bmatrix}
        \right )  
    \end{aligned}    
\end{equation}

As shown above, $\bm{\Sigma}_{t, t+1}$ is the cross-covariance computed through $2p$ dimensional unscented transformation $g'$ which resembles $g$. We proceed by calculating conditional distribution of random variable $\btheta_t | \btheta_{t+1}, \cD_T$. Notice that $\btheta_t$ is conditionally independent from $\{y_{t+1}, y_{t+1}, \cdots, y_{T}\}$ given $\btheta_{t+1}$. Therefore, we have

\[
\bP(\btheta_t | \btheta_{t+1}, \cD_T) = \bP(\btheta_t | \btheta_{t+1}, \cD_t)
\]

Thus, the desired conditional distribution can be calculated via Equation \ref{eq:trackCrossCov}. 
Due to the fact all distributions above are Gaussian, the conditional distribution can be calculated directly using Multivariate Normal property. Therefore, we have

\begin{align*}
    \btheta_{t} | \btheta_{t+1}, \cD_T &\sim \rN(\bm{m}'_t, \bm{C}'_t) \\
    \bm{D}_t &= \bm{\Sigma}_{t, t+1}(\Xi_{\bm{\Sigma}}(\btheta_t) + \bm{W})^{-1}\\
    \bm{m}_t' &= \bm{m}_t + \bm{D}_t[\btheta_{t+1} - \Xi_{\bm{\mu}}(\btheta)]\\
    \bm{C}_t' &= \btheta_t - \bm{D}_t (\Xi_{\bm{\Sigma}}(\btheta_t) + \bm{W}) \bm{D}_t^T
\end{align*}

As we've assumed that we know $\btheta_{t} | \cD_T \sim \rN(\bm{m}^*_t, \bm{C}^*_t)$, multiply $\bP(\btheta_{t+1} | \cD_T)$ with the above conditional distribution will result in $\bP(\btheta_t, \btheta_{t+1} | \cD_T)$. Writing in the matrix representation, the final joint distribution of $\btheta_t, \btheta_{t+1} | \cD_T$ can be calculated as below:

\begin{align*}
    \left. \begin{bmatrix}
    \btheta_t \\
    \btheta_{t+1}
    \end{bmatrix} \right | \cD_T &\sim \rN \left(
    \begin{bmatrix}
    \bm{m}_t^* \\
    \bm{m}_{t+1}^*
    \end{bmatrix},
    \begin{bmatrix}
    \bm{C}_t^* & \bm{\Sigma}_{t, t+1}^* \\
    {\bm{\Sigma}_{t+1, t}^*}^T & \bm{C}_{t+1}^*
    \end{bmatrix}
    \right )
    \\
    \begin{bmatrix}
    \bm{m}_t^* \\
    \bm{m}_{t+1}^*
    \end{bmatrix} &= 
    \begin{bmatrix}
    \bm{m}_t + \bm{D}_t [\bm{m}_{t+1}^* - \bm{m}_t] \\
    \bm{m}_{t+1}^*
    \end{bmatrix}
    \\
    \begin{bmatrix}
    \bm{C}_t^* & \bm{\Sigma}_{t, t+1}^* \\
    {\bm{\Sigma}_{t+1, t}^*}^T & \bm{C}_{t+1}^*
    \end{bmatrix} &=
    \begin{bmatrix}
    \bm{D}_t\bm{C}_{t+1}^*\bm{D}_t^T + \bm{C}'_{t+1} & \bm{D}_t\bm{C}_{t+1}^* \\
    \bm{C}_{t+1}^*\bm{D}_t^T & \bm{C}_{t+1}^*
    \end{bmatrix}\\
    \btheta_t | \cD_T &\sim \rN(\bm{m}_t^*, \bm{C}_t^*)\\
    \bm{m}_t^* &= \bm{m}_t + \bm{D}_t [\bm{m}_{t+1}^* - \bm{m}_t]\\
    \bm{C}_t^* &= \bm{D}_t\bm{C}_{t+1}^*\bm{D}_t^T + \bm{C}'_{t+1}
\end{align*}

The last step is the marginal distribution extracted from this joint distribution. If we start from $t+1 = T$, where $\btheta_{t+1} | \cD_{T} = \btheta_{T} | \cD_{T}$ is available, repeating the above steps recursively can end up constructing the following incomplete matrix as defined in Equation \ref{eq:jointBigMatrix}. In this following matrix, parameters in the mean vector and on the covariance matrix diagonal will be sufficient for performing backward smoothing. However, only having marginal smoothing covariance matrices for each time step is not sufficient for parameter inference, which will be discussed in Section \ref{sec:Param_Inference}. All parameters defined in this section including Equation \ref{eq:jointBigMatrix} stays the same for future sections. We will revisit all these defined parameters in Equation \ref{eq:jointBigMatrix} in the later parameter inference Section \ref{sec:Param_Inference}, in which smoothed cross-covariance matrices $\bm{\Sigma}_{t, t+1}^*$ will be further revisited for calculating optimization loss function gradients.

\begingroup
\renewcommand*{\arraystretch}{1.5}
\setlength\arraycolsep{2pt}
\begin{equation}\label{eq:jointBigMatrix}
    \bm{\Theta}  | \cD_T = \left.
    \begin{bmatrix}
    \btheta_0\\
    \btheta_1\\
    \btheta_2\\
    \btheta_3\\
    \vdots\\
    \btheta_{T-1}\\
    \btheta_{T}\\
    \end{bmatrix} \right | \cD_T
    \sim \rN \left(
    \begin{bmatrix}
    \bm{m}_{0} \\
    \bm{m}_{1} \\
    \bm{m}_{2} \\
    \bm{m}_{3} \\
    \vdots \\
    \bm{m}_{T-1} \\
    \bm{m}_{T} 
    \end{bmatrix} 
    ,
    \begin{bmatrix}
    \bm{C}_0^* & \bm{\Sigma}_{0, 1}^* & \quad & \quad & \cdots & \quad & \quad\\
    \bm{\Sigma}_{1, 0}^* & \bm{C}_1^* & \bm{\Sigma}_{1, 2}^* & \quad &  \cdots & \quad & \quad\\
    \quad & \bm{\Sigma}_{2, 1}^* & \bm{C}_2^* & \bm{\Sigma}_{2, 3}^* & \cdots & \quad & \quad\\
    \quad & \quad & \bm{\Sigma}_{3, 2}^* & \bm{C}_3^* & \cdots & \quad & \quad \\
    \vdots & \vdots & \vdots & \vdots & \ddots & \vdots &  \vdots \\
    \quad & \quad & \quad & \quad & \cdots & \bm{C}_{T-1}^* & \bm{\Sigma}_{T-1, T}^* \\
    \quad & \quad & \quad & \quad & \cdots & \bm{\Sigma}_{T, T-1}^* & \bm{C}_T^* \\
    \end{bmatrix} \right)     
\end{equation}
\endgroup

Above is a detailed model formulation. An intuitive model illustration can be found below in Figure \ref{fig:Model_Illustration}.

\begin{figure}[ht]
    \begin{center}
    \includegraphics[width = \linewidth] {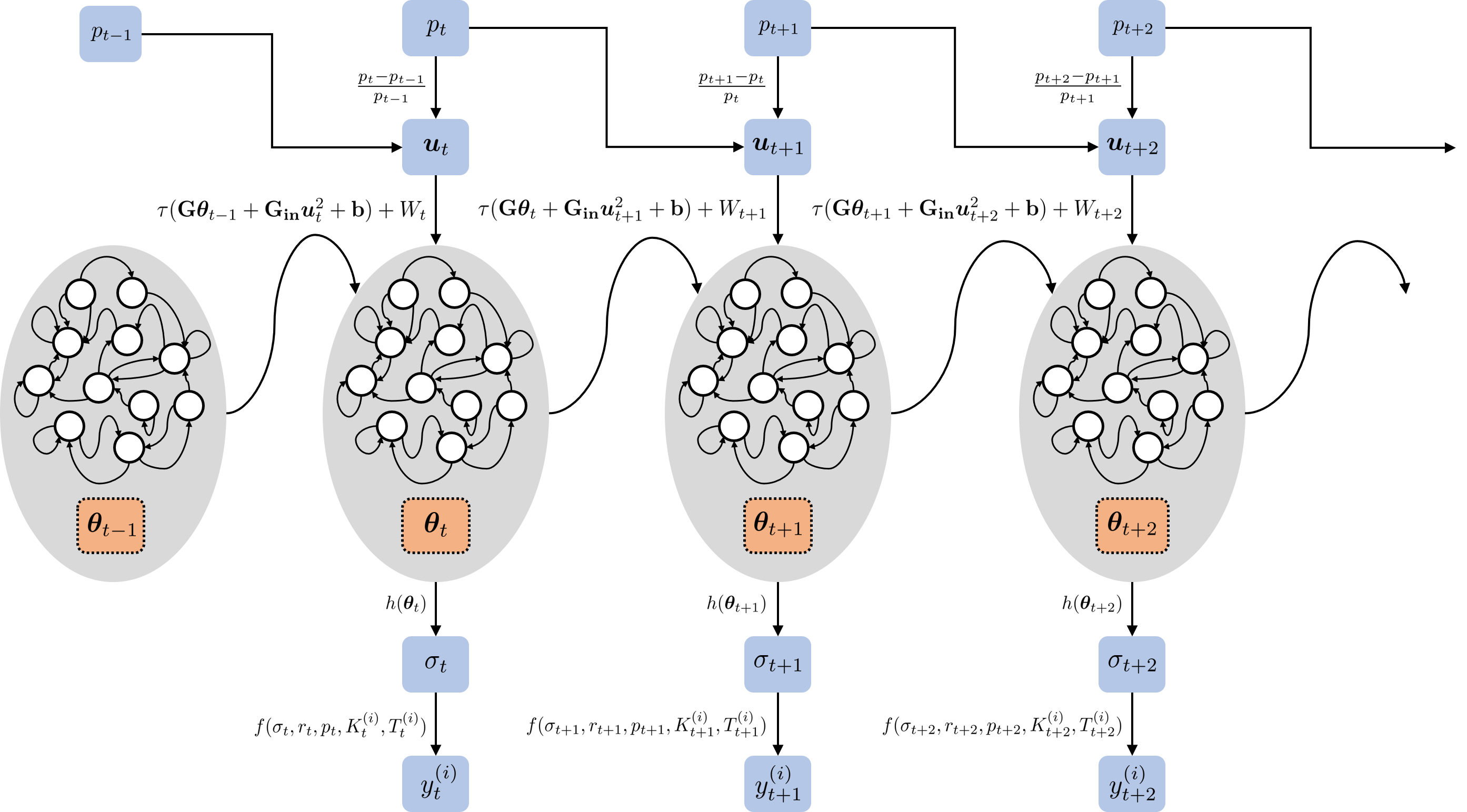}
    \end{center}
    \vspace{0.1in}
    \figurecaption{Figure \ref{fig:Model_Illustration}}{Model structure illustration of unscented reservoir smoother. Parameters included in blue boxes represents deterministic values whereas parameters included in orange boxes represent random variable. In our case, the random variable $\btheta$ is a stochastic dynamic modeled jointly by ESN and UKF.}  
    \captionlistentry{}
    \label{fig:Model_Illustration}
\end{figure}

\section{Model Parameter Inference} \label{sec:Param_Inference}

In the Kalman filter framework, the latent dynamic is constantly updated and predicted using Bayesian update from observation $\cD_t$. Such evolution function from $\btheta_t$ to $\btheta_{t+1}$ is fixed and pre-specified, in which the parameters are not updated or optimized throughout time. Though under ideal circumstances, the predetermined evolution function does capture the temporal patterns of the time series. However, when temporal pattern is complex and unknown, only by adapting the latent state distribution $\btheta_t |\cD_t$ will result in poor forecast and non-informative posterior distribution $\btheta_t | \cD_t$ and smoothed distribution $\btheta_t | \cD_T$. Here, though the evolution function $g$ lies in the framework of reservoir computing, serious overfitting may occur while increasing the size of the reservoir $p$. Therefore, it is necessary to make inference on the evolution function $g$ simultaneously while updating $\btheta_t$.

In this section, we will provide 2 different model parameter inference methods: The first method is an offline training algorithm that relies on the generalized EM algorithm for parameter optimization. Due to the fact this is essentially a gradient based optimization problem, it is possible to apply proper regularization to penalize expected likelihood, which is the main reason this method out-perform the second method. The second method was initially proposed by \citep{Wan2000}, which is an online training algorithm that casts both the parameters $\{ \bG, \bG_{in}, \bias\}$ of unscented Kalman filter and its latent state vectors $\btheta_t$ into a flattened vector for joint UKF update as defined previously in Section \ref{sec:FF}. This online algorithm doesn't rely on optimization, but rather self-update both the weight and latent state estimates using Bayesian Update. Due to this reason, regularization is not available, therefore leading to a weaker performance than the first algorithm. Derivation and description of both models will be further explained in Section \ref{sec:Offline} and \ref{sec:Online}.

\subsection{Offline Parameter Inference: Generalized EM} \label{sec:Offline}

In this section we explain making inference on parameters using generalized EM (GEM) algorithm. There are multiple ways of training unscented Kalman filter equipped with neural network as evolution function. MCMC, a Bayesian inference method, is also well known for providing posterior distribution inference for the parameters. However, we stick to training using EM due to several reasons. First, generalized EM algorithm is essentially an optimization problem. Thus, penalization could be directly applied on parameter values to obtain shrinkage effect. Second, different from general nonlinear evolution function, reservoir computing method introduces vast amount of weight parameters. The high dimensionality of parameters will make MCMC mixing mode difficult and slower to achieve. Besides, prior distribution for matrix valued parameters are more sophisticated to design. Thus, GEM provides relatively faster inference and more flexibility in parameter inference procedure.

EM algorithm is a technique for parameter inference based on observable data. The latent states $\bTheta := \btheta_{1:T}$, since not observable, are treated as missing data in this case. The objective of EM algorithm is to maximize the expected conditional log-likelihood of observable data $\cD_T$ given a set of not observable latent random variables and model parameters $\cG := \{\bG, \bG_{in}, \bias, \bW, v \} \in \bm{\Omega}$

\[
\cG = \argmax_{\cG \in \bm{\Omega} }  \bE_{\bTheta | \cG, \cD_{T}}[ \log \mathcal{L} ( \cD_T , \bTheta; \cG)]
\]

The challenge is that the closed form marginal conditional distribution $\bP( \bTheta |  \cG, \cD_T )$ is not available. EM algorithm \citep{Dempster1977} functions as an iterative method that seeks for local maximum of the above objective function by switching from pre-initialized $\cG_j$ to $\cG_{j+1}$, which maximizes expected likelihood defined below. This iterative method is composed of 2 steps: the E-step and M-step:

\begin{itemize}
    \item \textbf{E-step}: Instead of directly finding $\arg\max_{\cG}  \bE_{\bTheta | \cG}[ \log \mathcal{L} ( \cD_T , \bTheta ; \cG)]$, EM switch from $\cG_j$ to $\cG_{j+1}$ by
    
    \[
        \cG_{j+1} = \argmax_{\cG \in \bm{\Omega}}  \bE_{\bTheta | \cD_T, \cG_j}[ \log \mathcal{L} ( \cD_T , \bTheta ; \cG)]
    \]
    
    Therefore, we need to calculate the log-likelihood  $\log \mathcal{L} ( \cD_T, \bTheta; \cG) = \log \bP( \cD_T , \bTheta; \cG)$. It is a function of $\cD_T , \bTheta, \cG$, and we will integrate $\bTheta$ out via distribution $\bP( \bTheta | \cD_T, \cG_j)$, which will result into a function of $\cD_T$ and $\cG$. 
    
    \item \textbf{M-step}: As long as expected log-likelihood is calculated, EM algorithm finds the targeting parameter $\cG \in \bm{\Omega}$ that obtains global maximum of the expected log-likelihood. Under most circumstances, a closed form solution cannot be achieved. But the mechanism of EM, in theory, only requires updating $\cG$ that increases the expected log likelihood. Therefore, Generalized EM algorithm provides an alternative that only requires parameter $\cG_{j+1} $ to increases the expected log-likelihood comparing to $\cG_j$. This is sufficient to guarantee finding local maxima. Usually, GEM optimizes by gradient based methods.
\end{itemize}

EM algorithm iterates over E-step and M-step until a local maxima is obtained. Below, we will dive into derivation of E-step and M-step in our context. By direct calculation can we get expression for log-likelihood 

\begin{align*}
    \mathcal{L}( \cD_T , \bTheta; \cG) &= \log \bP( \cD_T , \bTheta; \cG )\\
    &= \sum_{t=1}^T \left\{  \log \bP(\btheta_t | \btheta_{t-1}, \cD_T) +  \sum_{i=1}^{I}\log (\bP(y_{t}^{(i)} | \btheta_t, \cD_T)  \right\} 
\end{align*}

Where we're only interested in the top $I$ most traded options on each date. Then, by expanding likelihood, we end up with non-linear transformation of random variable. Conditional distribution $\btheta_{t+1} | \btheta_t ,\cD_t$ could be again approximated via unscented transform as defined in notation Table \ref{table:notation}, where 

\begin{align*}
    \btheta_{t+1} = g(\btheta_t) \approx \Xi(\btheta_t) &\sim \rN(\Xi_\mu(\btheta_t), \Xi_\Sigma (\btheta_t))\\
    y_{t}^{(i)} = (f^{(i)} \circ h) (\btheta_t) \approx \Psi(\btheta_t)  &\sim \cN(\Psi_\mu^{(i)}(\btheta_t), \Psi_\Sigma^{(i)} (\btheta_t)) 
\end{align*}

Therefore, we can approximate the log-likelihood by

\begin{equation} \label{eq:loglikelihood}
    \begin{aligned}
        \mathcal{L}( \cD_T , \bTheta; \cG) & = -\frac{T}{2}(\log(|\bW|) - \frac{TI}{2}(\log(v)) \\
        & \quad -\frac{1}{2}\sum_{t=1}^{T}\left[\btheta_t^T\bW^{-1}\btheta_t - 2\btheta_t^T\bW^{-1}\Xi_\mu(\btheta_{t-1}) +\Xi_\mu(\btheta_{t-1})^T \bW^{-1}\Xi_\mu(\btheta_{t-1}) \right] \\
        & \quad -\frac{1}{2v}\sum_{t=1}^{T}[\sum_{i=1}^{I}(y_{t}^{(i)})^2 - 2\sum_{i=1}^{I} y_{t}^{(i)}\Psi_\mu^{(i)}(\btheta_t) + \sum_{i=1}^{I}(\Psi_\mu^{(i)}(\btheta_t))^2]  + C  
    \end{aligned}
\end{equation}

Equation \ref{eq:loglikelihood} is the log-likelihood of the entire process. See Appendix \ref{apx:derivation_5.1}. Then, we're taking expectation of Equation \ref{eq:loglikelihood} via conditional probability $\bP( \bTheta | \cD_T, \cG_j)$ to get expected log-likelihood:

\begin{equation}\label{eq:Expectedloglikelihood}
\begin{aligned}
&\mathbb{E}_{\bTheta \mid \cD_T, \cG_j}\!\left[\log \mathcal{L}(\cD_t, \bTheta \mid \cG)\right]\\
=\, &-\frac{T}{2}\Bigg(
\log|\bW|
- \frac{TI}{2}\log(v)
-\frac{1}{2}\sum_{t=1}^{T}\Big\{
\underbrace{\mathbb{E}\!\left[\btheta_t^T\bW^{-1}\btheta_t \mid \cD_T\right]}_{\text{(i)}}
-2\underbrace{\mathbb{E}\!\left[\btheta_t^T\bW^{-1}\Xi_\mu(\btheta_{t-1}) \mid \cD_T\right]}_{\text{(ii)}}
+\underbrace{\mathbb{E}\!\left[\Xi_\mu(\btheta_{t-1})^T \bW^{-1}\Xi_\mu(\btheta_{t-1})\mid \cD_T\right]}_{\text{(iii)}}
\Big\}\\
&\qquad\qquad
-\frac{1}{2v}\sum_{t=1}^{T}\Big\{
\sum_{i=1}^{I}(y_{t}^{(i)})^2
-2\sum_{i=1}^{I} y_{t}^{(i)}\underbrace{\mathbb{E}\!\left[\Psi_\mu^{(i)}(\btheta_t)\mid \cD_T\right]}_{\text{(iv)}}
+\sum_{i=1}^{I}\underbrace{\mathbb{E}\!\left[\Psi_\mu^{(i)}(\btheta_t)^2\mid \cD_T\right]}_{\text{(v)}}
\Big\}
\Bigg)
+ C .
\end{aligned}
\end{equation}

Equation \ref{eq:Expectedloglikelihood} is the expected log-likelihood that we want to obtain maximum by picking the optimal $\cG$. For detailed derivation, see Appendix \ref{apx:derivation_5.2}. Clearly, we cannot find the closed form global maxima for this expression. However, we may use constrained gradient based optimization methods, such as projected Newton (\citet{Bertsekas1982}, \citet{Kim2010}) or proximal quasi-Newton methods (\citet{Patriksson1998}, \citet{Lee2014}), to achieve a better set of parameters $\cG_{j+1}$.  

Moreover, due to non-linearity, there are 5 non-closed-form expressions within Equation \ref{eq:Expectedloglikelihood}. To use gradient based method, these terms need to be in analytical expressions. Such expressions are listed as below, in which all parameter definitions can be found in model formulation Section \ref{sec:Model_Formulation}. 

\begin{enumerate}
    \item[\textbf{i}]   $\bE [ \btheta_t^T \bW^{-1} \btheta_t | \cD_T ] = \text{tr}( \bW^{-1}\bC_t^*) + (\bm{m}_t^*)^T( \bW)^{-1}\bm{m}_t^*$. This is obtained using the standard expectation of quadratic form.
    \item[\textbf{ii}]  $\bE[ \btheta_t^T \bW^{-1} \Xi_\mu( \btheta_{t-1})  |\cD_T]$ is computation intractable. Details regarding to solving computational intractability of this term will be discussed later in Section \ref{sec:JointUT} and \ref{sec:Taylor}.
    \item[\textbf{iii}] $\bE[ \Xi_\mu( \btheta_{t-1})^T  \bW^{-1} \Xi_\mu( \btheta_{t-1}) |\cD_T] = \text{tr}( \bW^{-1} \Xi_{\Sigma}( \btheta_{t-1}  |\cD_T)) + \Xi_{\mu}( \btheta_{t-1}  |\cD_T)^T( \bW)^{-1} \Xi_{\mu}( \btheta_{t-1}  |\cD_T)$. Similar to \textbf{i}, this is obtained by expectation of quadratic form.
    \item[\textbf{iv}]  $\bE[\Psi_\mu^{(i)} ( \btheta_t)  |\cD_T] = \Psi_{\mu}^{(i)}(\bE[ \btheta_t |\cD_T])$. This is obtained by applying unscented transform on a random variable.
    \item[\textbf{v}]   $\bE[\Psi_\mu^{(i)}( \btheta_t)^2  |\cD_T] = [\Psi_{\mu}^{(i)}(\bE[ \btheta_t |\cD_T])]^2 + \Psi_{\Sigma}^{(i)}(\bE[ \btheta_t |\cD_T])$
\end{enumerate}

The second term are more difficult to estimate due to the fact that cross-covariance of $\btheta_{t-1}$ and $\btheta_t$ are intractable without closed form specification of the linearized evolution function $g$. We provide the following 2 ways to tackle this problem: joint unscented transformation and Taylor linearization.

\subsubsection{Joint Unscented Transform} \label{sec:JointUT}

The first way to compute $\bE[ \btheta_t^T \bW^{-1} \Xi_\mu( \btheta_{t-1})  |\cD_T] = \bE[*]$ is to approximate by unscented transform. The major challenge of estimating the cross covariance is that the nonlinear regression breaks the tractability of cross covariance between $\btheta_t, \btheta_{t-1}$. Therefore, it is necessary to design a transformation on the joint random variable $(\btheta_t, \btheta_{t-1}) | \cD_T$ in order to maintain the tractability of off-diagonal cross-covariance. Following this motivation, similar to the RTS smoothing method that has been described in Section \ref{sec:Back_Smoothing} , we could concatenate $\btheta_{t-1} | \cD_t$ and $\btheta_{t}  | \cD_t$ into a $2p$ dimensional joint random vector variable, and define joint transformation function $g''$ to represent original transformation $g$. 

\begin{align*}
    g''\left(\left. \begin{bmatrix}
    \btheta_{t-1}\\
    \btheta_t \\
    \end{bmatrix} \right )\right| \cD_T &= 
    \left(\left. \begin{bmatrix}
    g(\btheta_{t-1})\\
    \btheta_t \\
    \end{bmatrix} \right )\right| \cD_T \approx 
    \left. 
    \begin{bmatrix}
    \Xi(\btheta_{t-1})\\
    \btheta_t \\
    \end{bmatrix} \right| \cD_T 
    \\
    g'' &= 
    \begin{bmatrix}
    g \\
    I \\
    \end{bmatrix}
\end{align*}

As above, applying unscented transformation to approximate $g''$ and can keep track of $\btheta_t |\cD_T$ and $\Xi(\btheta_{t-1} | \cD_T)$. To further preserve approximation fidelity, $g''$ can be further decomposed as composite of linear transformation and also logistic evolution function $\tau$. As the linearly transformed random variable can be expressed exactly, we could use unscented transform to approximate only the $\tau$. The linear transform step is as follow:

\begin{align*}
    \left. \begin{bmatrix}
    \btheta_{t-1}\\
    \btheta_t
    \end{bmatrix} \right\vert \cD_t
    &\sim  \rN \left(
    \begin{bmatrix}
    \bm{m}_{t-1}^*\\
    \bm{m}_t^*
    \end{bmatrix}
    ,
    \begin{bmatrix}
    \bC_{t-1}^* & \bm{\Sigma}_{t-1, t}^* \\
    \bm{\Sigma}_{t, t-1}^* & \bC_t^*\\
    \end{bmatrix}
    \right)
    \\
    \left. \begin{bmatrix}
    \bG \btheta_{t-1} + \bG_{in} \bu_t^2 + \bias\\
    \btheta_t\\
    \end{bmatrix} \right\vert \cD_t
    & \sim  \rN \left(
    \begin{bmatrix}
    \bG \bm{m}_{t-1}^* + \bG_{in}\bu_t^2 + \bias\\
    \bm{m}_t^*\\
    \end{bmatrix}
    ,
    \begin{bmatrix}
    \bG \bC_{t-1}^*\bG^T & \bG \bm{\Sigma}_{t-1, t}^* \\
    \bm{\Sigma}_{t, t-1}^*\bG^T & \bC_t^* \\
    \end{bmatrix}
    \right)
\end{align*}

Then, apply unscented transform to approximate $\tau$. This 2 step approximation can further to preserve fidelity of the overall transformation $g''$. After that, take the off diagonal matrix $\bm{\Sigma}^{**}_{t,t-1} $ as cross covariance matrix to calculate $\bE[ \btheta_t^T \bW^{-1} \Xi_\mu( \btheta_{t-1})  |\cD_T]$

\begin{align*}
    \left. 
    \begin{bmatrix}
    \btheta_t \\
    \Xi(\btheta_{t-1})\\
    \end{bmatrix} \right| \cD_T &\sim 
    \rN \left(
    \begin{bmatrix}
    \bm{m}_t^*\\
    \Xi_{\bm{\mu}}(\btheta_{t-1})\\
    \end{bmatrix}
    ,
    \begin{bmatrix}
    \bC_t^* & \bm{\Sigma}^{**}_{t,t-1} \\
    \bm{\Sigma}^{**}_{t-1,t} & \Xi_{\Sigma}(\btheta_{t-1}) \\
    \end{bmatrix}
    \right)
    \\
    \bE[ \btheta_t^T \bW^{-1} \Xi_\mu( \btheta_{t-1})  |\cD_T] &= \text{tr}(\bm{\Sigma}^{**}_{t,t-1}) + (\bm{m}_t^*)^T    \Xi_{\bm{\mu}}(\btheta_{t-1})
\end{align*}

Thus, the numeric result of the second expectation term can be obtained as above.

\subsubsection{Taylor Linearization} \label{sec:Taylor}

Another way to solve intractability problem is to use Taylor linearization, where considerable amount of estimation fidelity are sacrificed in exchange for better computation efficiency. Due to the fact we're already computing quadratic form, the highest degree we can get for Taylor linearization is degree 1, as higher degree Taylor expansion will incorporate higher order terms whose expectations are again unable to compute. By RTS smoother, we are able to get $\bP(\btheta_t, \btheta_{t-1} | \cD_T)$, the joint distribution of 2 Gaussian random variables. Therefore,
\[
\bP(\btheta_t, \bG \btheta_{t-1} + \bG_{in}\bu_t^2+\bias | \cD_T)
\]
This is still Gaussian and exact distribution is known. We then want to get
\[
\bE(\btheta_t^T \bW^{-1} \tau(\bG \btheta_{t-1} + \bG_{in}\bu_t^2) +\bias | \cD_T)
\]
Notice that we are using $\tau$ as logistic function. Therefore $\tau(\bG \btheta_{t-1} + \bG_{in} \bu_t^2 + \bias)$ is logit-normal distribution \citep{Atchison1980}, whose distribution has a skewed bell-shaped when the standard deviation its logit is small. Therefore, by expanding at smoothed posterior mean $\bm{m}_t^*$, we can obtain that

\[
    \bE[ \btheta_t^T \bW^{-1} \Xi_\mu( \btheta_{t-1})  |\cD_T] \approx (\bm{m}_{t}^*)^T \bW^{-1} \tau(\bG \bm{m}_{t-1}^* + M) + N \cdot \mathrm{diag}[\bW^{-1}\bm{\Sigma}_{t, t-1}^*\bG ^T]
\]

Where

\begin{align*}
    \eta(x_0) &:= \frac{e^{-x_0}}{(1 + e^{-x_0})^2}\\
    M &= \bG _{in} \bu_t^2 + \bias\\
    N &= \eta(\bG \bm{m}_{t-1}^* + M)
\end{align*}

This expression is in closed form and is easy to compute. Thus, the Taylor linearization approach returns faster results comparing to the joint unscented transform in Section \ref{sec:JointUT}. For detail derivation of Taylor linearization, see Appendix \ref{apx:Taylor} for details.

\subsubsection{Forming Offline Loss Function} \label{sec:Regularization}

Above are 2 ways to obtain the second conditional expectation term introduced in Equation \ref{eq:Expectedloglikelihood}. Choosing between them is the trade-off between computation efficiency and approximation fidelity. Unscented transform, due to its need to compute transformation for each sigma points, usually results in slower computation. But higher precision is guaranteed. On the other hand, Taylor linearization is faster, however can cause more significant estimation error, especially when the Logit normal distribution has bigger Logit standard deviation. To clarify, all further analysis within this article are based on using the unscented  method for nonlinear approximation, as we're more interested in the precision of the model.

As all 5 expectation terms can be computed, Equation \ref{eq:Expectedloglikelihood} becomes part of the loss function the algorithm will be optimizing. Besides, in addition to maximizing the expected log-likelihood, we can regularize parameters of $\bG$ and $\bG_{in}$. As previously described in Section \ref{sec:Modification_to_Model}, $\bias$ is introduced in the model only as a mean-shift factor. It serves to self-adjust and to find a proper mean-shift of the evolution function in order to stabilize the state space model, which doesn't impact the complexity of the model. Therefore, no regularization is placed upon $\bias$. Finally, for potential model interpretability concerns, we'd hope to increase sparsity within weights of $\bG, \bG_{in}$. Therefore, we'd regularize $\bG, \bG_{in}$ via Lasso $L_1$ loss \citep{Tibshirani1996} to promote sparsity within the dynamical system. 

The finalized offline training loss function is in Equation \ref{eq:lossFunc}, where $\alpha$ is a hyper-parameter coefficient adjusting weight of regularization. A detailed outline of offline learning algorithm is defined in Algorithm \ref{algo:Offline}.

\begin{align}
    \mathcal{R}_{\text{em}}(\cG, \cD_T) &= \text{Equation } \ref{eq:Expectedloglikelihood} \nonumber \\
    \mathcal{R}_{\text{lasso}}(\cG) &=  \| \bG \|_{L_{1}} + \| \bG_{in} \|_{L_{1}} \nonumber \\
    \mathcal{R}(\cG, \cD_T) &= \mathcal{R}_{\text{em}}(\cG) + \alpha \mathcal{R}_{\text{lasso}}(\cG, \cD_T) \label{eq:lossFunc}
\end{align}

{
\vspace{0.25in}
\begin{algorithm}[h]
\caption{Offline parameter inference}
\label{algo:Offline}
\SetAlgoLined
\KwResult{ A full Smoothed trajectory $\bTheta$ of latent state $\btheta_t$, parameter sets including $\{ \bG, \bG_{in}, \bias, \bW, v \}$ }
 \textbf{Initialize}: $\{\bG, \bG_{in}, \bias \}$ by Algorithm 1; $\bW, v$, $\btheta_0 \sim \rN( \bm{m}_0, \bC_0)$ \\
 \Repeat{Validation set error is minimized}
 {
    \For{ $ t \in \{1, 2, \cdots T\}$  }
    {   
        \begin{itemize}
            \itemsep0em
            \item Approximate $\btheta_t | \cD_{t-1} \sim \rN(\bm{a}_t , \bm{R}_t)$ by unscented  as 
            \[
                \btheta_t | \cD_{t-1} \sim \rN(\Xi_{\mu}[\btheta_{t-1}|\cD_{t-1}] , \Xi_{\Sigma}[\btheta_{t-1}|\cD_{t-1}]) = \rN(\bm{a}_t , \bm{R}_t)
            \]
            \item Calculate $\btheta_t | \cD_t \sim \rN(\bm{m}_t , \bC_t)$ as Bayesian update of forward filtering.
        \end{itemize}
    }
    \For{ $ t \in \{T, T-1, \cdots 1\}$} 
    {
        \begin{itemize}
            \itemsep0em
            \item Calculate $( \btheta_{t-1}, \btheta_t)^T | \cD_T$ recursively by RTS backward smoothing. 
            \item Record: $\btheta_{t-1} | \cD_T \sim \rN(\bm{m}_{t-1}^* , \bC_{t-1}^*)$, and also cross-covariance $\bm{\Sigma}_{t-1,t}^*$.
        \end{itemize}
    }
    \textbf{Calculate}: EM loss function as in Equation \ref{eq:Expectedloglikelihood} and Lasso regularization loss. Then calculate loss function $\mathcal{R}(\cG, \cD_T) $ as in \ref{eq:lossFunc}.\\
    \textbf{Calculate}: $\nabla \mathcal{R}(\cG, \cD_T)$ and optimize $ \cG := \{\bG, \bG_{in}, \bias, \bW, v \} $
 }
\textbf{Return}: $\bG, \bG_{in}, \bias$
\end{algorithm}
\vspace{0.25in}
}

\subsection{Online Parameter Inference: Joint UKF} \label{sec:Online}

Inspired by \citet{Wan2000}, we could filter both the state estimates $\btheta_t$ and parameters $\bG, \bG_{in}, \bias$ jointly. Rather than treating $\bG, \bG_{in}, \bias$ as a fixed predetermined parameter that determines the entire $\bTheta$, we could again cast parameters $\bG, \bG_{in}, \bias$ into the Kalman filter framework. 

Consider $\cG^{flat}$ as the concatenation of flattened $\cG' := \{ \bG, \bG_{in}, \bias \}$.
\[
\cG^{flat} := [\bG^{flat}, \bG_{in}^{flat}, \bias^{flat}]^T
\]

Notice that covariance $\bW$ and variance $v$ are not part of the parameters for training. As we assume that the parameters $\cG'$ stays the same $\forall t$, equivalent to the previous model defined in Equation \ref{eq:trackCrossCov}, we define identity transition function over the parameter:

\[
\cG^{flat}_{t+1} = \cG^{flat}_{t}
\]

In this way, we can construct a joint unscented Kalman filter equivalent to previous model as below

\begin{align*}
    \bTheta_{\cG}^{(t+1)} =
    \begin{bmatrix}
    \btheta_{t+1} \\
    \cG^{flat}_{t+1} 
    \end{bmatrix}
    &=
    g^{+}\left(
    \begin{bmatrix}
    \btheta_{t} \\
    \cG^{flat}_{t} 
    \end{bmatrix}
    \right) + \bW^+
    = 
    \begin{bmatrix}
    g(\btheta_{t}, \cG^{flat}_{t} ) \\
    I(\cG^{flat}_{t}) 
    \end{bmatrix} + \bW^+\\
    \bTheta_{\cG}^{(t)} &\sim \rN(\bm{m}_{t}^+, \bC_{t}^+)
\end{align*}

Where $\cG_{t}^{flat}$ are parameters that parametrizes the original transition function $g$. During forward filtering steps, we're updating both distribution for parameter sets and latent state vectors.  Notice that at each iteration, we allow an variance on $\cG_{t}^{flat}$. This modification endows the parameter set $\cG_t$ the time-varying ability, therefore allowing the parameter to re-adapt to the time series if needed. After forward filtering, similarly, we can apply RTS backward smoothing on the the concatenated vector, which returns a complete smoothed trajectory of distribution of both $\btheta_t$ and $\cG^{flat}$ that traces back to $t=0$, which we set a prior upon.

Then, we can set the newly smoothed posterior of $(\btheta_0, \cG^{flat}_0)^T$ as the new prior and repeat the forward filtering and backward smoothing steps described above. Since the above processes allows weights and state estimate to change by performing Kalman update with a continuous in-stream of data, this is essentially an online training algorithm, that $y_{T+1}$ will update not only $\btheta_{T+1} | \cD_{T+1}$, but also update $\bG, \bG_{in}, \bias$ when making forecasts. The formal description of the algorithm is included at Algorithm \ref{algo:Online}.

{
\vspace{0.25in}
\begin{algorithm}[H]
\caption{Online parameter inference}
\label{algo:Online}
\SetAlgoLined
\KwResult{ A full Smoothed trajectory $\bTheta$ of latent state $\btheta_t$, and a full trajectory of $\cG^{flat} := \{\bG, \bG_{in}, \bias\}$}
 \textbf{Initialize}: $ \{\bG, \bG_{in}, \bias \}$ by Algorithm 1; $\bW, v$; $\btheta_0 = \bm{m}_0$. Then flatten parameters to get $\cG^{flat} := \{\bG, \bG_{in}, \bias \}$ and concatenate $(\btheta_0, \cG^{flat} )^T = (\btheta_0, \bG, \bG_{in}, \bias)^T$. Then initialize covariance matrix for concatenated vector $\bTheta_{\cG}^{(0)} \sim \rN( \bm{m}_0^+, \bm{C}_0^+)$\\
\Repeat{Validation set error is minimized}
 {
    \For{ $ t \in \{1, 2, \cdots T\}$  }
    {   
        \begin{itemize}
            \itemsep0em
            \item Approximate $ \bTheta_{\cG}^{(t)} | \cD_{t-1} \sim \rN(\bm{a}_t^+ , \bm{R}_t^+)$ by unscented .
            \item Calculate $\bTheta_{\cG}^{(t)} | \cD_t \sim \rN(\bm{m}_t^+ , \bC_t^+)$ as Bayesian update of forward filtering.
        \end{itemize}
    }
    \For{ $ t \in \{T, T-1, \cdots 1\}$} 
    {
        \begin{itemize}
            \itemsep0em
            \item Calculate $( \bTheta_{\cG}^{(t-1)}, \bTheta_{\cG}^{(t)})^T | \cD_T$ recursively by RTS-backward smoothing. 
            \item Record: $\bTheta_{\cG}^{(t-1)} | \cD_T \sim \rN({\bm{m}_{t-1}^+}^* , {\bC_{t-1}^+}^*)$
        \end{itemize}
    }
    \textbf{Reset}: $\bTheta_{\cG}^{(0)} \sim \rN({\bm{m}_{0}^+} , {\bC_{0}^+}) =  \rN({\bm{m}_{0}^+}^* , {\bC_{0}^+}^*)$
 }
\textbf{Return}: $\bTheta_{\cG} := (\btheta_t, \bG_{(t)}, {\bG_{in}}_{(t)}, \bias_{(t)})^T \forall t \in \{0,1,2 \cdots, T\}$ the full trajectory of latent state vector $\bTheta$ and reservoir parameters $\cG'$ .
\end{algorithm}
\vspace{0.25in}
}

This algorithm, when making prediction, is an online algorithm. Comparing to the offline training algorithm doesn't require calculating EM loss. Therefore it is faster and more convenient for training. However, due to the inability to include regularization, the algorithm usually result in a less satisfactory result than the offline learning algorithm. To achieve similar performance, the reservoir size $p$ is considerably smaller than that of the offline learning framework, and therefore leading to an even faster learning speed.

\section{Simulations} \label{sec:Simulation}

This section evaluates the URS's performance on forecasting point estimates and uncertainty quantification. To address this, we generate synthetic data using Cox–Ingersoll–Ross (CIR) process \cite{Cox1985}, which is used by Heston model to model the rough stochastic volatility process. The SPDE form can be expressed as

\[
d V_{t}=\theta \left(\mu - V_{t}\right) d t+ \sigma \sqrt{V_{t}} d W_{t}
\]

Where $V_t$ denotes the volatility rough process measured in variance, $\mu$ denotes the long-term volatility, $\theta$ measures the converging rate of volatility to the $\mu$, $\sigma$ is the volatility (standard deviation) of the volatility process, and $W_t$ is a Brownian motion. The CIR process exhibit the behaviour that converges to the long-term volatility $\mu$. The process' parameters have interpretable meanings, making the simulation justifiable. Besides, as we will be comparing results of our model URS and GARCH \citep{Bollerslev1986}, HARCH \citep{Muller1997} family, it is reasonable to generate simulated data that satisfies the competitors', GARCH and HARCH, assumptions, as they are essentially special discrete cases of the generalized CIR process. To best test URS's capability, for all the CIR process we generated below, we generated by setting $\sigma = 0.04$ and $\theta = 10$. $\sigma = 0.04$ makes the ground truth volatility process $V_{0:t}$ a noisy process, which aligns with the real market behaviour. Other parameters specification are case-specific. The code that we have used to such CIR process samples is from a public available GitHub repository created by \citet{JasonAsh2020}  \footnote{
Pyesg Github repository link \url{https://github.com/jason-ash/pyesg}} .

Essentially, the goal is to simulate option data. Once a volatility trajectory $V_{0:t}$ has been sampled, we use $V_{0:t}$ process as building blocks to generate synthetic option price data. Considering that each economist has estimation error for $V_{t}$, we add independent yet identically distributed irreducible error $\epsilon^{(ir)}_t \sim N(0,\kappa_V)$ to $V_{0:t}$ to form $I$ deviated volatility $V_t^{(1)}, V_t^{(2)}, \cdots, V_t^{(I)}$ for each time step, and each of them will contribute 1 observed option price. The complete synthetic data generation procedure is summarize as in Algorithm \ref{algo:SyntheticData}.

{
\vspace{0.25in}
\begin{algorithm}[h] 
\caption{Synthetic data generation}
\label{algo:SyntheticData}
\SetAlgoLined
\KwResult{1 set of simulated option data}
\textbf{Initialize}: $n$: length of the volatility process; $V_0$: the initial value of volatility (measured in standard deviation) process; $\mu$: the long-term limiting volatility of the ground truth volatility process; $p_0$ initial price; $\kappa_V$: the irreducible error added to ground truth volatility process; $K$: the number of observed options on each date.\\
\textbf{Generate}: ground truth volatility process following Cox–Ingersoll–Ross (CIR) process $V_{1:n}$ with standard deviation of volatility $\sigma = 0.04$
\[
d V_{t}=\theta \left(\mu - V_{t}\right) d t+ \sigma \sqrt{V_{t}} d W_{t}
\]
\textbf{Generate}: $k$ deviated volatility processes by adding random noise following N$(0, \kappa_V)$ at each time step $V_t$, forming $\{V_{t}^{(1)}, V_{t}^{(2)}, \cdots, V_{t}^{(I)}\}\forall t$
\[
u_t \sim N(0, V_t)  \quad  \forall t
\]
\textbf{Calculate}: observed price process $p_t$
\[
p_t = p_{t-1}(u_{t}+1)  \quad  \forall t>0
\]
\textbf{Calculate}: $y_{t}^{(i)}$ option prices for the $i^{th}$ option at time $t$ using random expiration date $T_{t}^{(i)}$, random strike price $K_{t}^{(i)}$, and $p_t$, ${V_t}^{(i)}$, and risk free rate $r_t = 0.02$.  \\
\textbf{Return}: $u_t, p_t, y_t, T_{t}^{(i)}, K_{t}^{(i)}$ for modeling purposes, and $V_t$ process for ground truth comparison.
\end{algorithm}
\vspace{0.25in}
}

By generating synthetic data, we explore both coverage probability and the point estimate performance of unscented reservoir smoother. Performance has been explored under stationary scenarios and non-stationary scenarios.

For stationary scenario, using algorithm 4, we generated 10 set of simulated data with $\kappa_V = 0.01$, $n=200$, $\sigma_0 = \mu = 0.15$, and $p_0 = 2000$. By measuring theoretical coverage versus the observed coverage for prediction step from 1 to 20, we observe that URS tends to exhibit better coverage behaviour for larger $k$ step predictions. This is due to the fact that the logistic mapping function $\tau$ is a contraction mapping with a bounded range. To sustain posterior predictive distribution variance not from degenerating, innovation variance $\bm{W}$ and posterior distribution $\btheta_t | \cD_t$ need to compensate by increase uncertainty. Thus, a flatter predictive priori distribution $\btheta_{t+1} | \cD_t$ is obtained. Besides, observable output $y_{t}^{(i)}$ are generated based on deviated volatility process $V_{t}^{(i)}$, which is obtained adding the additional irreducible error $\kappa_V$ to ground truth, the predictive distribution variance is thus a result of adding the theoretical irreducible estimate uncertainty $\kappa_V / \sqrt{I}$ to the modeling induced uncertainty. 

Below in Figure \ref{fig:stationarySimulation} (a), we present smoothed (blue) state confidence interval, predicted state uncertainty interval (red), and theoretical irreducible estimate uncertainty $\kappa_V / \sqrt{I}$ (yellow). We observe that both the smoothed and predicted state confidence interval successfully cover the entire irreducible estimate uncertainty confidence interval, showing that URS uncertainty quantification doesn't underestimate the irreducible sample error. To obtain modeling induced uncertainty, we minus total uncertainty by the theoretical irreducible estimate uncertainty $\kappa_V / \sqrt{I}$. In Figure \ref{fig:stationarySimulation} (b), we observe that not too much additional error was induced by the modeling process. 

\begin{figure}[hbtp]
    \begin{center}
    \begin{subfigure}[b]{1\textwidth}
        \centering
        \includegraphics[width = \linewidth]{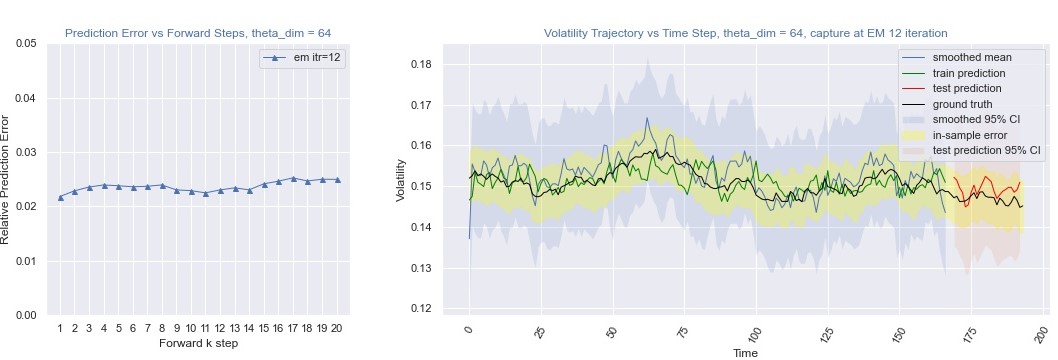}
        \figurecenter{(a) Uncertainty with in-sampled error $\kappa_V / \sqrt{I}$}
    \end{subfigure}
    \begin{subfigure}[b]{1\textwidth}
        \centering
        \includegraphics[width = \linewidth]{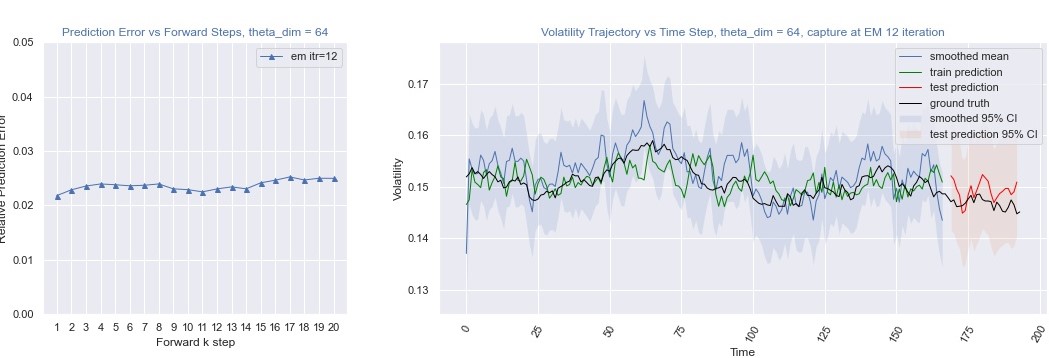}
        \figurecenter{(b) After removal of in-sample error $\kappa_V / \sqrt{I}$: modeling induced uncertainty}
    \end{subfigure}
    \end{center}
\figurecaption{Figure \ref{fig:stationarySimulation}}{Predictive distribution of stationary synthetic series. The series is generated via Cox-Ingersoll-Ross with initial volatility not equals its asymptotic converging limit. The smoothed means of training set $\btheta_{1:T}$ are plotted blue line with their $95\%$ credible interval estimate in shaded blue. 1 step prediction of training set is plotted in green line. The 1 step forecast of testing set is plotted in red line with its $95\%$ interval in shaded red. Ground truth is plotted in solid black line. The left column indicates the $k$ step prediction error. URS tends to do equally well in both shorter-term and longer-term forecast when non-stationary trend doesn't present in the time series.
(a) The yellow shade indicates the $95\%$ irreducible estimate uncertainty $\kappa_V / \sqrt{I}$. The blue and red shade completely covers the yellow shade, indicating that URS does not underestimate noise within the data. Furthermore, the forecasting mean lies entirely in the yellow shade, indicating URS' superior point estimate.
(b) The blue shade is calculated by subtracting $\kappa_V / \sqrt{I}$ from the original credible interval of $\btheta_{1:T}$, indicating the extra modeling induced uncertainty. Its width is narrow, demonstrating URS' superior uncertainty measurement.}
\captionlistentry{}
\label{fig:stationarySimulation}
\end{figure}

Similarly, for non-stationary scenario, we also generated 10 set of simulated data with $\kappa_V = 0.01$, $n=200$, $\sigma_0 = 0.2, \mu = 0.15$, and $p_0 = 2000$. Under non-stationarity, we observe that URS is capable of capturing trend due to outside control $u_t$. Similarly, the posterior variance is big. However, we can verify by observing that the in-sample error $\kappa_V / \sqrt{I}$ takes much proportion of contribution to total uncertainty. Below in Figure \ref{fig:nonstationarySimulation} (a), we present smoothed (blue) state confidence interval, predicted state uncertainty interval (red), and theoretical irreducible estimate uncertainty $\kappa_V / \sqrt{I}$ (yellow). We observe that both the smoothed and predicted state confidence interval successfully cover the entire irreducible estimate uncertainty confidence interval, showing that URS uncertainty quantification doesn't underestimate the irreducible sample error. Minus total uncertainty by the theoretical irreducible estimate uncertainty $\kappa_V / \sqrt{I}$, we get only modeling induced uncertainty In Figure \ref{fig:nonstationarySimulation} (b).

\begin{figure}[hbtp]
    \begin{center}
        \begin{subfigure}[b]{1\textwidth}
        \centering
        \includegraphics[width = \linewidth]{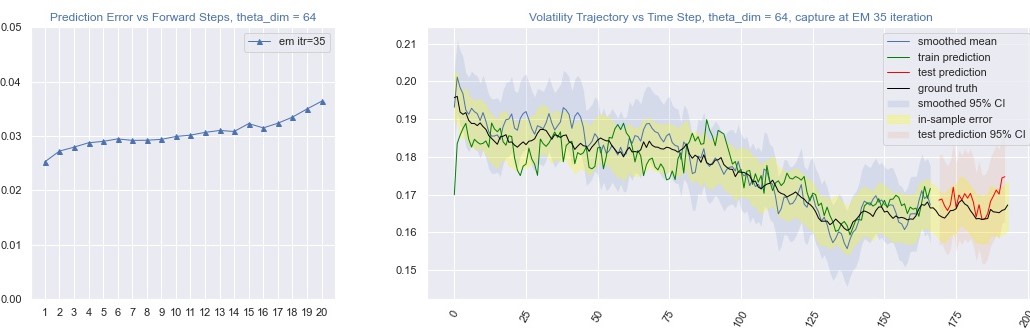}
        \figurecenter{(a) Uncertainty with in-sampled error $\kappa_V / \sqrt{I}$}
    \end{subfigure}
    \begin{subfigure}[b]{1\textwidth}
        \centering
        \includegraphics[width = \linewidth]{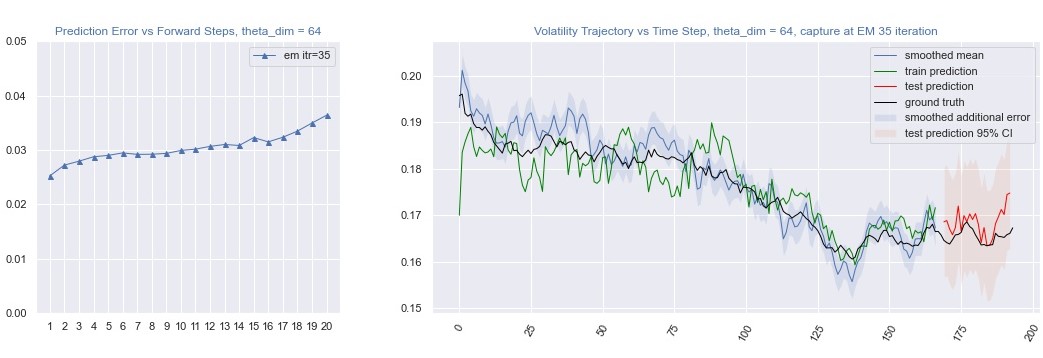}
        \figurecenter{(b) After removal of in-sample error $\kappa_V / \sqrt{I}$: modeling induced uncertainty}
    \end{subfigure}
    \end{center}
\figurecaption{Figure \ref{fig:nonstationarySimulation}}{Predictive distribution under non stationary series. The series is generated via Cox-Ingersoll-Ross with initial volatility not equals its asymptotic converging limit. The smoothed means of training set $\btheta_{1:T}$ are plotted blue line with their $95\%$ credible interval estimate in shaded blue. 1 step prediction of training set is plotted in green line. The 1 step forecast of testing set is plotted in red line with its $95\%$ interval in shaded red. Ground truth is plotted in solid black line. The left column indicates the $k$ step prediction error. URS tends to do better in longer-term forecast when non-stationary trend presents in the time series. 
(a) The yellow shade indicates the $95\%$ irreducible estimate uncertainty $\kappa_V / \sqrt{I}$. The blue and red shade completely covers the yellow shade, indicating that URS doesn't underestimate noise within the data. Furthermore, the forecasting mean lies entirely in the yellow shade, indicating URS' superior point estimate.  
(b) The blue shade is calculated by subtracting $\kappa_V / \sqrt{I}$ from the original credible interval of $\btheta_{1:T}$, indicating the extra modeling induced uncertainty. Its width is still reasonable, demonstrating URS' valid uncertainty measurement.}
\captionlistentry{}
\label{fig:nonstationarySimulation}
\end{figure}

\begin{figure}[hbtp]
    \begin{center}
        \begin{subfigure}[b]{0.49\textwidth}
            \centering
            \includegraphics[width=\textwidth]{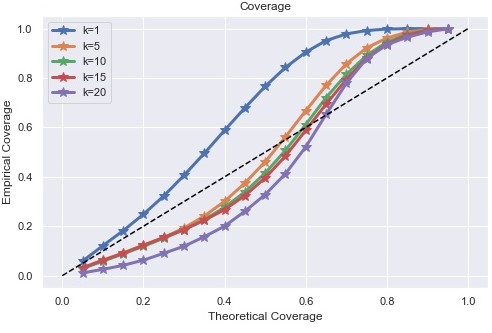}
            \figurecenter{(a) Stationary (URS)}
        \end{subfigure}
        \begin{subfigure}[b]{0.49\textwidth}
            \centering
            \includegraphics[width=\textwidth]{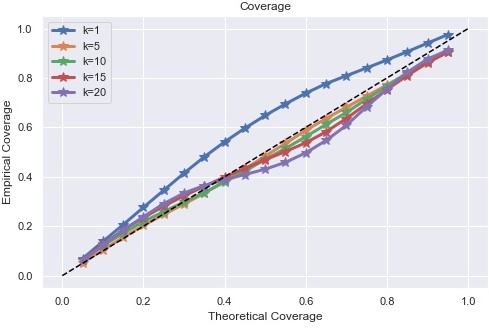}
            \figurecenter{(d) Non-stationary (URS)}
        \end{subfigure}
        
        \begin{subfigure}[b]{0.49\textwidth}
            \centering
            \includegraphics[width=\textwidth]{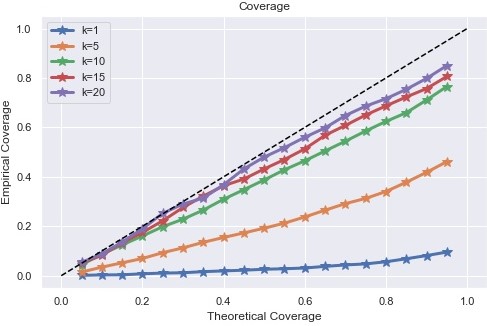}
            \figurecenter{(b) Stationary (GARCH)}
        \end{subfigure}
        \begin{subfigure}[b]{0.49\textwidth}
            \centering
            \includegraphics[width=\textwidth]{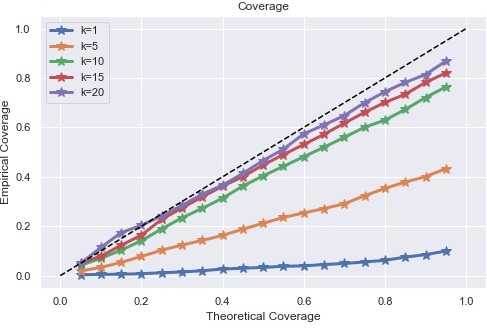}
            \figurecenter{(e) Non-stationary (GARCH)}
        \end{subfigure}
        
        \begin{subfigure}[b]{0.49\textwidth}
            \centering
            \includegraphics[width=\textwidth]{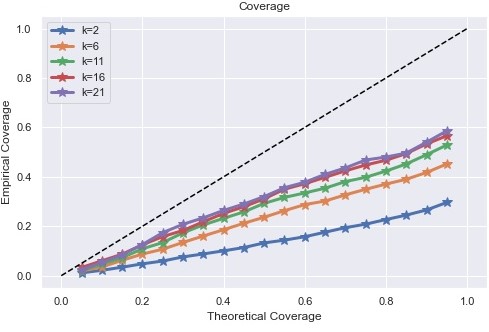}
            \figurecenter{(c) Stationary (HARCH)}
        \end{subfigure}
        \begin{subfigure}[b]{0.49\textwidth}
            \centering
            \includegraphics[width=\textwidth]{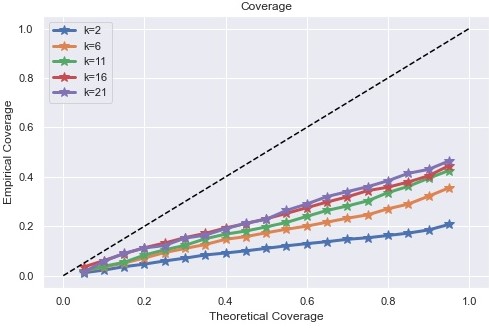}
            \figurecenter{(f) Non-stationary (HARCH)}
        \end{subfigure}
    \end{center}
\figurecaption{Figure \ref{fig:coverage}}{Coverage Probability. The coverage probability is obtained by calculating percentage of forecasting credible intervals that covers the ground truth. For a $\alpha$ percentage credible interval, we expect $\alpha$ percent of predicted intervals to contains the ground truth, Thus, the closer the line is to the diagonal line (dotted black line), the better the uncertainty measurement. (a), (b), (c) compares coverage for stationary series. URS obviously outperform the its competitors. However, URS tends to produce larger interval estimates. (d), (e), (f) compares coverage for stationary series. URS still outperform the its competitors. (a) and (d) demonstrate that when trend exists, URS' more conservative (wider) interval estimates increases uncertainty measurement.}
\captionlistentry{}
\label{fig:coverage}
\end{figure}

In Figure \ref{fig:coverage} we explored the observed coverage probability versus the theoretical coverage to evaluate uncertainty measurement of URS comparing to other stochastic volatility models. The first phenomenon is that URS tends to make comparatively wider interval estimates for shorter-term predictions. There are 2 reasons causing this phenomenon, one would be that $\tau$ is bounded evolution function. Therefore, shorter-term interval estimates are made wider to counter degenerative contraction evolution function. Another is that the predictive distribution contains theoretical irreducible uncertainty which has been discussed above. Thus, shorter-term coverage tends to be higher than expected. Another phenomenon is that URS exhibit better uncertainty measurement for non-stationary process than the stationary process. Generally, URS obviously outperformed in uncertainty measurement comparing to both GARCH and HARCH. Furthermore, in Table \ref{table:Synthetic_Performance}, URS outperformed in point estimates. To further compare point estimates of URS with that of more models, including deep LSTM \citep{Hochreiter1997}, that doesn't offer interval estimates, we proceed into testing model comparison on real world data, which is discussed in Section \ref{sec:Experiment}.

\begin{table}[H]
\resizebox{\textwidth}{!}{
\begin{tabular}{cccccccccccc}
\Xhline{2\arrayrulewidth}
\multirow{2}{*}{Data}           & \multirow{2}{*}{Model} & \multicolumn{5}{c}{Relative Prediction Error} & \multicolumn{5}{c}{ $95 \%$ Coverage} \\ \cline{3-12} 
                                &                        & $k=1$  & $k=5$  & $k=10$ & $k=15$ & $k=20$ & $k=1$  & $k=5$  & $k=10$ & $k=15$ & $k=20$  \\ \hline
\multirow{4}{*}{Stationary}     & \textbf{URS}           & \textbf{0.0248} & \textbf{0.0280} & \textbf{0.0283} & \textbf{0.0295} & 0.0302 & 
                                                        \textbf{1.0000} & \textbf{0.9996} & \textbf{0.9996} & \textbf{0.9991} & \textbf{0.9989}  \\
                                & GARCH                  & 0.0612 & 0.0573 & 0.0523 & 0.0514 & 0.0526 & 0.0954 & 0.4615 & 0.7653 & 0.8060 & 0.8500  \\
                                & HARCH                  & 0.0330 & 0.0308 & 0.0302 & 0.0299 & \textbf{0.0292} & 0.2963 & 0.4520 & 0.5287 & 0.5660 & 0.5860  \\ \hline
\multirow{4}{*}{Non-stationary} & \textbf{URS}           & \textbf{0.0262} & \textbf{0.0313} & \textbf{0.0312} & \textbf{0.0320} & \textbf{0.0319} & 
                                                        \textbf{0.9757} & \textbf{0.9045} & \textbf{0.9071} & \textbf{0.9050} & \textbf{0.9122}  \\
                                & GARCH                  & 0.0689 & 0.0648 & 0.0595 & 0.0592 & 0.0609 & 0.0996 & 0.4330 & 0.7640 & 0.8200 & 0.8680  \\
                                & HARCH                  & 0.0459 & 0.0459 & 0.0464 & 0.0473 & 0.0476 & 0.2075 & 0.3565 & 0.4253 & 0.4540 & 0.4640  \\ \Xhline{2\arrayrulewidth}
\end{tabular}}
\vspace{10pt}
\caption{\small Synthetic data point estimate and coverage probability comparison result}
\label{table:Synthetic_Performance}
\end{table}

\section{Experiments} \label{sec:Experiment}

This section implements the 2 parameter inference algorithms (offline and online) together with some competitor algorithms to demonstrate the behaviour and performance of URS in forecasting real-world data. Training was guided by comparing performance of training set and validation set in a slightly different way which will be described later in this section.

\subsection{Data Preparation} \label{sec:Data_Prep}

The dataset used in the empirical analysis are the Standard and Poor 500 index option from the beginning of year 2015 to the end of 2019. The source of the dataset is from a standard research database, \textit{Wharton Research Data Services (wrds)} \footnote{wrds link \url{https://wrds-www.wharton.upenn.edu/}} \citep{WhartonSchoolattheUniversityofPennsylvania2013}. The database is comprehensive. It records daily option quotes for both call and put options. Due to our interest, we filtered only call options. For each option quote, there are a number of different recorded expiration dates $T_{t}^{(i)}$, and for each quote with a specified expiration dates, there are several options tickers with different strike prices $K_{t}^{(i)}$. A unique ticker with a defined $T_{t}^{(i)}, K_t^{(i)}$ defines one row of the dataset. Each row also records the best (highest) bid price of the option and the best offer (lowest) price of the option. These 2 prices sets a reasonable range of strike price $K_{t}^{(i)}$ in which the option deal could happen. In order to achieve a single value ground truth, we take the average of the 2 prices and use the mean as the ground truth of option price for further analysis. 

In our experiment, we pick the top $I = 5$ mostly traded option for each date. This is important because options prices can be very high or very low depending on expiration date $T$ and strike price $K$. Thus, it is very possible that the option price is very far away from the truth. By picking the most traded options according to their traded volumes will at least filter out most of the unreasonable ones, therefore ensuring the correctness of our fitting data $\cD_T$.  

Besides that, we also need two other datasets: one is a standard SP500 historical price dataset \footnote{SPX500 dataset link: \url{https://finance.yahoo.com/quote/\%5ESPX/history/} } \citep{Wankel2012}, which provides the asset price time series, $p_t$. Another is the London InterBank Offer Rate (LIBOR) 3-month rate for U.S. dollars \footnote{LIBOR dataset link: \url{https://fred.stlouisfed.org/series/USD3MTD156N}} \citep{ICEBenchmarkAdministrationLimited2015}. We will use this dataset as the risk-free interest rate series, $r_t$, in the Black-Scholes formula. 

The option data is broken into 3 parts: training part, validation part, and testing part. Different from the usual situation where the model is trained on some sample time series and making prediction on other sample time series, unscented reservoir smoother trains and forecasts on a single series time series. Therefore, the break up of training, validation, and testing are sequential in time, where training part comes first, followed by a validation part, and then testing part. Notice that the longer the validation, the better we are informed about choice of the hyper-parameter. However, too long a validation set will push testing set further back, which leads to a worse and inaccurate tested model performance. Therefore, we generally only assigned 1 or 2 time steps for validation set, merely used to probe the choice of the hyper-parameters and serves as a guidance to stop the algorithm, and an additional approximately 25 time steps for testing set, while the rest of the series prior to the validation set are all used for training purposes.

\subsection{Model Comparison} \label{sec:Model_Comp}

Unscented reservoir Smoother lays in the interface of both sequential deep learning models and State-space models. Therefore, we're interested in model performance comparing to both related stochastic volatility models and some deep learning models. A list of stochastic volatility model we've incorporated are Generalized Autoregressive Conditional Heteroskedasticity (GARCH) and HARCH, which is an extension from GARCH. On the other hand, the list of deep learning sequential models we will compare with are deep LSTM, and LSTM with Black-Scholes Formula as the output function. The latter is designed to exclude the ``burden'' of LSTM in approximating the already known output function, and thus makes the model comparison more objective. Finally, we have also incorporate implied volatility from option data as an additional comparison.

The wrds provided dataset already contains a field denoting the implied volatility based on market observation on that date. However, in real world, there is a volatility smile (calibration) that cannot be capture by Black-Scholes formula. This will lead to a systematic underestimate or overestimate of the implied volatility and the output option price that even we know the correct volatility, the output of Black-Scholes formula with the correct volatility doesn't match the option price. Therefore, merely by transforming the implied volatility using Black-Scholes into option price will result in a very poor prediction. In order to achieve a more reasonable result, we're also providing a calibrated implied volatility via inverse function that is computed numerically

\[
\sigma_t = f^{-1}(y_t)
\]

This returns the implied volatility assuming that market makers used exactly the Black-Scholes formula to price the option. Thus, the calculated implied volatility leads to the best possible prediction result. This is crucial as model comparison should compare the result of URS with the best possible results produced by its competitors.

Similarly, for GARCH and HARCH model, we also calibrate the predicted volatility process by adding the best possible constant to achieve their best possible performance.

For LSTM, the deep learning competitor, as the model is trained by directly fitting the option data, no prediction calibration is needed. Besides, We also compare results of LSTM when its output function has been clearly pre-stated as the Black-Scholes formula, hoping that such concatenation can further improve prediction.

\subsection{Testing Methodology} \label{sec:Test_Method}

The testing performance metric is the relative prediction error, defined as below, where Equation \ref{eq:single_test_loss} defines error generated from a predictions of a single step and \ref{eq:overall_test_loss} defines the overall average error.

\begin{align}
    \epsilon_{k \text{-step}}^{(t)} &= \frac{1}{I}\sum_{i}^{I}\frac{|\hat{y}_{t}^{(i)}- y_{t}^{(i)}|}{y_{t}^{(i)}} \label{eq:single_test_loss}\\
    \epsilon_{k \text{-step}} &= \mathrm{Mean}\left\{\epsilon_{k \text{-step}}^{(T+2)}, \epsilon_{k \text{-step}}^{(T+3)}, \cdots \right\} \label{eq:overall_test_loss}
\end{align}

Relative prediction error ensures smaller magnitude testing samples are not ignored and dominated. In this context, real world option data $y_{t}^{(i)}$ exhibits highly uneven magnitude. MSE is not informative as it conceals inefficiencies on testing samples of smaller magnitudes. Thus, relative prediction error ensures model performances on samples of all magnitudes are equally evaluated.

We study both short-term and long-term prediction performance. In this study, we compared $k=\{1, 2, \cdots , 20\}$ step performances. Forecasting predictions are generated via the following procedure:
 
\begin{enumerate}
    \itemsep0em
    \item Since filtered and smoother distribution of the $\btheta_T$ are identical, we take the filtered posterior distribution of the last state $\btheta_T | \cD_T$ in the training set as the ground truth.
    
    \item Set $\bP(\btheta_T | \cD_T)$ as prior distribution to extend prediction into future. Direct following the training set are the 2 time steps of observations in the validation set, denoted as $\bm{y}_{T+1},\bm{y}_{T+2}$. Validation errors are calculated by comparing ground truth $\bm{y}_{T+1} := (y_{T+1}^{(1)}, y_{T+1}^{(2)}, \cdots, y_{T+1}^{(I)})^T$ with predicted $\bE[\bm{y}_{T+1} | \cD_T] := (\hat{y}_{T+1}^{(1)}, \hat{y}_{T+1}^{(2)}, \cdots, \hat{y}_{T+1}^{(I)})^T$ and ground truth $\bm{y}_{T+1}$ with predicted $\bE[\bm{y}_{T+1} | \cD_T]$ via loss function defined in Equation \ref{eq:single_test_loss}. However, to forecast testing set, we perform Bayesian update at both $T+1$ and $T+2$ to obtain posterior distributions of $\btheta_{T+1} | \cD_{T+1}$ and $\btheta_{T+2} | \cD_{T+2}$. It is obvious that these 2 time steps of validation data did not contribute to inference of any model parameter $\cG$. Hence, the updated posterior distribution $\btheta_{T+2} | \cD_{T+2}$ is slightly different from ground truth. Due to this, the validation set has to be small and short to prevent large deviation of $\btheta_{T+2} | \cD_{T+2}$ from ground truth. Otherwise, the test set predictions, made available by extending forecast from the incorrect posterior $\btheta_{T+2} | \cD_{T+2}$, will be falsified.
    
    \item To test $k$ step prediction performance, we calculate distribution $\btheta_{T+2+k} | \cD_{T+2}$ from $\btheta_{T+2} | \cD_{T+2}$ without performing Bayesian update. Then, we calculate predictive mean $\bE[\bm{y}_{T+2+k} | \cD_{T+2}]$ calculate error by Equation \ref{eq:single_test_loss} with ground truth $\bm{y}_{T+2+k}$. This generates one $k$ step prediction error $\epsilon_{k \text{-step}}^{(t)}$ evaluated at $t = T+2$, which is temporarily saved.
    
    \item After completing the $k$ step testing at $t=T+2+k$, we perform Bayesian update and obtain the next ``ground truth'' $\btheta_{T+3} | \cD_{T+3}$ from $\btheta_{T+2} | \cD_{T+2}$. Then, by following step 3, we can generates another $k$ step prediction error $\epsilon_{k \text{-step}}^{(t)}$ evaluated at $t = T+3$, which is temporarily saved.
    
    \item Repeating step 3 and step 4 until the $k$-step forward reaches the last available time step of the entire testing set. In this process we have recorded $\{\epsilon_{k \text{-step}}^{(T+2)}, \epsilon_{k \text{-step}}^{(T+3)}, \cdots\}$. Averaging them results the overall $k$ step testing error. 
\end{enumerate}

\subsection{Result} \label{sec:Result}

Algorithms exhibit different behaviour provided with different length of training data. Hence, we run all comparable methods on both the longer time series (5 years) and the shorter time series (2 years). 

We first present results for the shorter data. To train URS offline, we set
$\btheta$ dimension equals to 8, and for online training, $\btheta$ dimension equals 4; The Lasso shrinkage coefficient $\alpha = 0.05$; $\bu$ dimension is 10; top $I$ number of options used is 5; Mean shift effect $\bias$ is initialized to be $-2.3$. The time series started from Jan $1^{st}$, 2018 and ended on Dec $31^{st}$, 2019. The last 24 days comprises the testing data. 1 day before testing set comprises the validation data, and the rest are training data. Result are shown in Table \ref{table:Model_Com}. Clearly, the offline URS achieves the best prediction results at all prediction steps. Even comparing with calibrated prediction for all stochastic volatility models or with the deep learning sequential model, its prediction is significantly superior. Besides, we notice that offline URS does comparably better prediction in long-term than in short-term. This is resulted from the RTS smoothing procedure that improved the algorithm's robustness to irreducible noise and also the ability to capture longer term trend. However, we observe that performance of the online algorithm for URS is far less effective.

We then present results for longer time series. To train URS offline, we set $\btheta$ dimension equal to 16, and we for online algorithm we still set $\btheta$ dimension equal to 4. The Lasso shrinkage coefficient was set to $\alpha = 0.05$; $\bu$ dimension is set to 10; top $I$ number of options used is 5; The mean shift effect $\bias$ is initialized to $-2.3$.  The time series started from Jan $1^{st}$, 2015 and ended on Dec $31^{st}$, 2019. The last 24 days comprises the testing data. 1 day before testing set comprises the validation data, and the rest are training data. Result are shown in the lower half of Table \ref{table:Model_Com}.

In contrast, though URS offline still outperform all other algorithms, we observe a significant performance improvement. The online algorithm typically doesn't perform as well given a short term series. However, when doubling the length of training set, its performance becomes comparable with calibrated stochastic volatility models. This is due to the nature of the algorithm's training mechanism, as longer time series provides more iterations for joint UKF updates. However, comparing to its offline counterpart, the online algorithm still lacks regularization. This provides intuition of why the online algorithm doesn't outperform the offline algorithm.

\begin{table}[H]
\small
\begin{center}
\begin{tabular}{llccccc}
\Xhline{2\arrayrulewidth}
\multicolumn{1}{c}{\multirow{2}{*}{Dataset}}                                                                                    & \multicolumn{1}{c}{\multirow{2}{*}{Model}} & \multicolumn{1}{l}{} & \multicolumn{3}{c}{Relative Prediction Error} & \multicolumn{1}{l}{} \\ \cline{4-6}
\multicolumn{1}{c}{}                                                                                                           & \multicolumn{1}{c}{}    & $k=1$        & $k=5$         & $k=10$        & $k=15$        & $k=20$        \\ \hline
\multirow{10}{*}{\begin{tabular}[c]{@{}l@{}}2-year Dataset\\ Jan 1st, 2018 \\ Dec 31st, 2019\end{tabular}} & Implied Volatility (IV) & 1.2422       & 4.2062        & 0.8061        & 0.7311        & 0.7267        \\
                                                                                                           & IV (calibrated)         & 0.3804       & 0.3400        & 0.3901        & 0.4422        & 0.4444        \\
                                                                                                           & GARCH                   & 0.5768       & 0.5953        & 0.6983        & 0.7049        & 0.6524        \\
                                                                                                           & GARCH (calibrated)      & 0.2821       & 0.2583        & 0.2043        & 0.2176        & 0.1957        \\
                                                                                                           & HARCH                   & 0.5261       & 0.5957        & 0.8018        & 0.8361        & 0.6239        \\
                                                                                                           & HARCH (calibrated)      & 0.3958       & 0.3886        & 0.4244        & 0.5085        & 0.3264        \\
                                                                                                           & LSTM                    & 0.7979       & 0.9817        & 0.8186        & 0.9934        & 0.9900        \\
                                                                                                           & LSTM (Black-Scholes)    & 1.0210       & 0.9045        & 1.3947        & 0.9790        & 0.9692        \\
                                                                                                           & \textbf{URS Offline}    & {\bf 0.2319} & {\bf 0.2149}  & {\bf 0.1698}  & {\bf 0.1574}  & {\bf 0.1081}  \\
                                                                                                           & URS Online              & 0.9036       & 0.8915        & 0.8763        & 0.8468        & 0.9375        \\ \hline
\multirow{10}{*}{\begin{tabular}[c]{@{}l@{}}5-year Dataset\\ Jan 1st, 2015 \\ Dec 31st, 2019\end{tabular}} & Implied Volatility (IV) & 1.2422       & 4.2062        & 0.8061        & 0. 7311       & 0.7267        \\
                                                                                                           & IV (calibrated)         & 0.3804       & 0.3400        & 0.3901        & 0. 4422       & 0.4444        \\
                                                                                                           & GARCH                   & 0.6096       & 0.5230        & 0.6918        & 0. 7076       & 0.5334        \\
                                                                                                           & GARCH (calibrated)      & 0.3743       & 0.3241        & 0.2479        & 0. 2205       & 0.3185        \\
                                                                                                           & HARCH                   & 0.4828       & 0.5300        & 0.6820        & 0. 6887       & 0.4571        \\
                                                                                                           & HARCH (calibrated)      & 0.4006       & 0.3825        & 0.3609        & 0. 4174       & 0.2192        \\
                                                                                                           & LSTM                    & 0.7687       & 0.9757        & 0.7305        & 0. 9934       & 0.9893        \\
                                                                                                           & LSTM (Black-Scholes)    & 0.8258       & 0.9598        & 0.9359        & 0. 9878       & 0.9809        \\
                                                                                                           & \textbf{URS Offline}    & {\bf 0.2212} & {\bf 0.2106}  & {\bf 0.1735}  & {\bf 0.1632} & {\bf 0.1157}  \\
                                                                                                           & URS Online              & 0.2833       & 0.2873        & 0.3025        & 0. 2954       & 0.3062        \\ \Xhline{2\arrayrulewidth}
\end{tabular}
\vspace{10pt}
\caption{\small Model prediction error comparison}
\label{table:Model_Com}
\end{center}
\end{table}

\section{Discussion} \label{sec:Discussion}

The former sections present comprehensive study of URS in the context of option pricing. However, the originality of unscented reservoir smoother is not limited to this single case. Due to the fact that URS lies in the intersection of deep learning and state-space model, it triggers our interest to explore further extensions of URS that has more profound implication on general problems. In this section, we discuss two levels of extensions. For simplicity purpose, notations and parameters in this section are not inherited from previous sections.

\subsection{Extensions on Control Input} \label{sec:Control_Extend}

A close observation of Equation \ref{eq:FinalEvolution} shows that the control input variable serves as an additive factor to the transition function. When applying unscented transform, such control only serves has a mean-shift effect on the state estimate distribution. Therefore, the input weight $\bG_{in}$ doesn't necessarily need to be linear. It could well be replaced by any functions that is differentiable. For example, input function can be a neural network with arbitrary structures. The input control is still an additive factor that doesn't impact the class of distribution of $\btheta_t$, which is still Gaussian. As for training and making inference on parameters, the offline parameter inference algorithm relies on calculating gradient to optimize E-step. But gradient based optimization is exactly the optimization regime for typical neural network training. Thus, the deep learning framework is fully compatible at the input control part of our framework. For example, consider an RNN extension based on the framework, we can extend the evolution function in Equation \ref{eq:FinalEvolution} to

\begin{align*}
    \btheta_t &= \tau(\bG\btheta_{t-1} + g_{in}(\bu_t) + \bias  ) + W_t\\
    W_t &\sim \rN(\bm{0}, \bW)
\end{align*}
    
The change of input control function doesn't modify the class of distribution of $\bm{h}_{t}$. Parameter inference is invariant and no additional approximation error is induced. 

A further extension would be to consider having multiple layers of recurrent structures before the control input. This includes deep ESN structures (\citet{Sun2017}, \citet{Chouikhi2018}, \citet{Ma2017}) or in any deep RNN structures with the last layer being ESN. In this case, the URS becomes

\begin{align*}
    \bm{h}_{t}^{(1)} &= \tau^{(1)}\left(\bG^{(1)} \bm{h}_{t-1}^{(1)} + g_{in}^{(1)}(\bm{x}_t)+ \bias^{(1)}\right)\\
    \bm{h}_{t}^{(2)} &= \tau^{(2)}\left(\bG^{(2)} \bm{h}_{t-1}^{(2)} + g_{in}^{(2)}(\bm{h}_t^{(1)}, \bm{x}_{t})+ \bias^{(2)}\right)\\
    &\vdots\\
    \bm{h}_{t}^{(n)} &= \tau^{(n)}\left(\bG^{(n)} \bm{h}_{t-1}^{(n)} + g_{in}^{(n)}(\bm{h}_t^{(n-1)}, \bm{x}_{t})+ \bias^{(n)}\right)\\
    \btheta_t &= \tau\left(\bG\btheta_{t-1} + g_{in}(\bm{h}_t^{(n)})  + \bias \right) + W_t\\
    W_t &\sim \rN(\bm{0}, \bW)
\end{align*}

Where $\bm{x}_t$ denotes input at $t$, $\bm{h}_{t}^{(i)}$ denotes the $i^{th}$ layer of recurrent unit at time $t$, $\bG^{(i)}$, $\bias^{(i)}$ denotes the transition weight and bias of the $i^{th}$ layer, and $t^{(i)}$ is the evolution function of the $i^{th}$ layer. In this case, simple algebra can show that in E-step, the expected log-likelihood is still a function of $\{\bm{h}_{1:T}^{(1:n)}, \bm{G}^{(1:n)}, \bias^{(1:n)}, \bm{x}_{1:T}\}$. Optimization is exactly the same as backpropagation through time. Besides, the probabilistic layer, the last UKF layer, still maintains additive control input. Thus, training no additional approximation error is induced. Therefore, we have demonstrated the full compatibility of deep learning and state space model on the input control level.

\subsection{At the Intersection of Deep Learning and State-Space Models} \label{sec:Full_Extend}

We make the final generalization of URS to integrate the sequential deep learning frameworks and the state-space model frameworks. The major challenge is that for more complex gated network structures, there are usually more than one dynamic whose uncertainty measurement needs to be propagated. To overcome this, we consider the generalization of a typical gated sequential deep learning module, LSTM.

\begin{align*}
\bm{f}_{t} &= \tau_{gate}^{(f)}\left( \bm{x}_{t}, \bm{h}_{t-1}  \right) \\
\bm{i}_{t} &= \tau_{gate}^{(i)}\left( \bm{x}_{t}, \bm{h}_{t-1}  \right) \\
\bm{o}_{t} &= \tau_{gate}^{(o)}\left( \bm{x}_{t}, \bm{h}_{t-1}  \right) \\
\tilde{\bm{c}}_{t} &= g_{cell}\left( \bm{x}_{t}, \bm{h}_{t-1} \right) \\
\bm{c}_{t} &= \bm{f}_{t} \circ \bm{c}_{t-1}+\bm{i}_{t} \circ \tilde{\bm{c}}_{t} \\
\bm{h}_{t} &= \bm{o}_{t} \circ f_{out}\left(\bm{c}_{t}\right)
\end{align*}

The existence of gates demonstrates the existence of more than one time varying states. In this LSTM scenario, the two dynamics whose uncertainty measurements need to be propagated are $\bm{c}_t$ and $\bm{h}_t$. This differs from all of our previous extensions where only one dynamic ($\btheta$) was cast into state-space model framework. For this type of multiple dynamics uncertainty propagation problem, a solution would be treating the all dynamics jointly as one concatenated dynamic. Thus, the above LSTM could also be cast into the state-space model framework as below

\begin{align*}
    \begin{bmatrix}
    \bm{c}_t\\
    \bm{h}_t
    \end{bmatrix}
    &= 
    g\left(
    \begin{bmatrix}
    \bm{c}_{t-1}\\
    \bm{h}_{t-1}
    \end{bmatrix}, 
    \bm{x}_t
    \right) + 
    \begin{bmatrix}
    W_t\\
    V_t
    \end{bmatrix}\\
    W_t &\sim \rN(\bm{0}, \bW) \\
    V_t &\sim \rN(\bm{0}, \bm{V})
\end{align*}

Where evolution function $g$ encapsulates all the gating and other composite of functions. Its explicit expression is
{\small
\[
g\left(
\begin{bmatrix}
\bm{c}_{t-1}\\
\bm{h}_{t-1}
\end{bmatrix}, 
\bm{x}_t
\right) =
\begin{bmatrix}
\tau_{gate}^{(f)}\left( \bm{x}_{t}, \bm{h}_{t-1} \right) \circ \bm{c}_{t-1}+ \tau_{gate}^{(i)}\left( \bm{x}_{t}, \bm{h}_{t-1} \right) \circ g_{cell}\left( \bm{x}_{t}, \bm{h}_{t-1} \right)\\
\tau_{gate}^{(o)}\left( \bm{x}_{t}, \bm{h}_{t-1} \right) \circ f_{out}\left(\tau_{gate}^{(f)}\left( \bm{x}_{t}, \bm{h}_{t-1} \right) \circ \bm{c}_{t-1}+ \tau_{gate}^{(i)}\left( \bm{x}_{t}, \bm{h}_{t-1} \right) \circ g_{cell}\left( \bm{x}_{t}, \bm{h}_{t-1} \right)\right)
\end{bmatrix}
\]
}
\normalsize

The general idea is to concatenate all time-varying states into a joint state, and the transformation function could be modeled via joint transformation. This extension holds for deep learning sequential models of fixed dynamic dimension in general, including model proposed by \citet{Doerr2018}. Although depending on the number of parameters, inference could switch from unscented Kalman filtering to SMC-type methods to achieve higher fidelity (see \citet{DelMoral1997} and \citet{Liu1998} ). Thus, the URS model motivates a complete integration of sequential deep learning framework and state-space model framework.

\section{Conclusions}  \label{sec:Conclusion}

Sequential models that lies in the deep learning framework, including reservoir computing, have demonstrated success in capturing complex sequential patterns of deterministic dynamical systems. However, we have demonstrated that these models suffers from learning from highly noisy data. Specifically, when observed output $\bm{y}$ contains large unobservable noise, the model performance degenerate. Besides, all time steps' induced prediction loss contribute by the same weight to the total loss function, so that essentially each training sample has equal power in affecting parameters. Nevertheless, some time steps' data samples contain larger unobservable irreducible noise $\epsilon^{(ir)}_t$ and therefore should contribute less in affecting model parameters, as their larger uncertainty indicates them being less informative or credible. This causes deep learning models' susceptibility to noise. Furthermore, their lack of a uncertainty quantification makes us unable measure and propagate predictive distribution. State-space models, on the other hand, provides measured distribution, allowing us to quantify the uncertainty. Thanks to this uncertainty measurement, information of model parameters drawn from observed outputs with higher irreducible uncertainty are therefore diluted, which improved the model's robustness to data noise. However, their lack of model complexity makes the framework not as flexible, leaving it harder to extract more complex temporal patterns from sophisticated time series. This study proposes the unscented reservoir smoother (URS) as a fundamental model and studies its behaviour in option pricing setting. This model lies in the intersection of both frameworks, capable of measuring uncertainty for both latent dynamics and outputs while simultaneously obtaining better, especially longer term, forecasting results. Most importantly, using URS as building blocks, we have demonstrated that further extensions could be made and therefore integration of sequential deep learning framework and state-space model framework is ultimately feasible.

% Acknowledgements should go at the end, before appendices and references

\acks{We would like to thank Duke Fuqua School of Business for providing us access to the Wharton Research Data Services (wrds). }

% Manual newpage inserted to improve layout of sample file - not
% needed in general before appendices/bibliography.

\newpage

\appendix
\begin{appendices}
\section{Derivations of Equations}

In this section we elaborate on derivations of computationally heavy equations appeared in this paper.

\subsection{Derivation of Equation \ref{eq:loglikelihood} } \label{apx:derivation_5.1}

\begin{align*}
    \mathcal{L}( \cD_T , \bTheta; \cG) & = \sum_{t=1}^T \left\{  \log \bP(\btheta_t | \btheta_{t-1}, \cD_T) +  \sum_{i=1}^{I}\log (\bP(y_{t}^{(i)} | \btheta_t, \cD_T)  \right\} \\
    & = \sum_{t=1}^{T} \left\{-\frac{1}{2}\log((2\pi)^p|\bW|)-\frac{1}{2}(\btheta_t - \Xi_\mu(\btheta_{t-1}) )^T \bW^{-1}(\btheta_t - \Xi_\mu(\btheta_{t-1}) ) \right.\\
    & \quad \left. -\frac{I}{2}\log(2\pi v) -\frac{1}{2v}\sum_{i=1}^{I}(y_{t}^{(i)} - \Psi_\mu^{(i)}(\btheta_t))^2\right\}  \\
    & =  -\frac{T}{2}(\log(|\bW|) - \frac{TI}{2}(\log(v)) \\
    & \quad -\frac{1}{2}\sum_{t=1}^{T}\left[\btheta_t^T\bW^{-1}\btheta_t - 2\btheta_t^T\bW^{-1}\Xi_\mu(\btheta_{t-1}) +\Xi_\mu(\btheta_{t-1})^T \bW^{-1}\Xi_\mu(\btheta_{t-1}) \right] \tag{\ref{eq:loglikelihood}}\\
    & \quad -\frac{1}{2v}\sum_{t=1}^{T}[\sum_{i=1}^{I}(y_{t}^{(i)})^2 - 2\sum_{i=1}^{I} y_{t}^{(i)}\Psi_\mu^{(i)}(\btheta_t) + \sum_{i=1}^{I}(\Psi_\mu^{(i)}(\btheta_t))^2]  + C 
\end{align*}

Equation \ref{eq:loglikelihood} is the full log-likelihood of the process. Then, we're taking expectation of Equation \ref{eq:loglikelihood} via conditional probability $\bP( \bTheta | \cD_T, \cG_j)$ to get expected log-likelihood:

\subsection{Derivation of Equation \ref{eq:Expectedloglikelihood} } \label{apx:derivation_5.2}

\begin{align}
\quad &\mathrm{E}_{\Theta \mid \mathcal{D}_T, \mathcal{G}_j }
\Big[ \log \mathcal{L}(\mathcal{D}_t, \Theta \mid \mathcal{G}) \Big]\\
=\, & \int_{\prod^T \mathcal{H}}
\log \mathcal{L}(\mathcal{D}_t, \Theta \mid \mathcal{G})\,
\mathrm{P}(\Theta \mid \mathcal{D}_T, \mathcal{G}_j)\, d\Theta \\
=\, & -\frac{T}{2}\Bigg(
\log|W| - \frac{TI}{2}\log v \nonumber \\
&\quad -\frac{1}{2}\sum_{t=1}^{T}\int_{\prod^T \mathcal{H}}
\mathrm{P}(\Theta \mid \mathcal{D}_T, \mathcal{G}_j)\,
\Big[
\theta_t^{\top}W^{-1}\theta_t
- 2\,\theta_t^{\top}W^{-1}\Xi_{\mu}(\theta_{t-1})
+ \Xi_{\mu}(\theta_{t-1})^{\top}W^{-1}\Xi_{\mu}(\theta_{t-1})
\Big]\, d\Theta \nonumber \\
&\quad -\frac{1}{2v}\sum_{t=1}^{T}\int_{\prod^T \mathcal{H}}
\mathrm{P}(\Theta \mid \mathcal{D}_T, \mathcal{G}_j)\,
\Big[
\sum_{i=1}^{I}\bigl(y_{t}^{(i)}\bigr)^2
- 2\sum_{i=1}^{I} y_{t}^{(i)}\,\Psi_{\mu}^{(i)}(\theta_t)
+ \sum_{i=1}^{I}\bigl(\Psi_{\mu}^{(i)}(\theta_t)\bigr)^2
\Big]\, d\Theta
+ C
\Bigg) \\
=\, & -\frac{T}{2}\Bigg(
\log|W| - \frac{TI}{2}\log v \nonumber \\
&\quad -\frac{1}{2}\sum_{t=1}^{T}\Big\{
\underbrace{\mathrm{E}\!\left[ \theta_t^{\top}W^{-1}\theta_t \,\middle|\, \mathcal{D}_T \right]}_{\text{(i)}}
- 2\underbrace{\mathrm{E}\!\left[ \theta_t^{\top}W^{-1}\Xi_{\mu}(\theta_{t-1}) \,\middle|\, \mathcal{D}_T \right]}_{\text{(ii)}}
+ \underbrace{\mathrm{E}\!\left[ \Xi_{\mu}(\theta_{t-1})^{\top}W^{-1}\Xi_{\mu}(\theta_{t-1}) \,\middle|\, \mathcal{D}_T \right]}_{\text{(iii)}}
\Big\} \nonumber \\
&\quad -\frac{1}{2v}\sum_{t=1}^{T}\Big\{
\sum_{i=1}^{I}\bigl(y_{t}^{(i)}\bigr)^2
- 2\sum_{i=1}^{I} y_{t}^{(i)}\,
\underbrace{\mathrm{E}\!\left[ \Psi_{\mu}^{(i)}(\theta_t) \,\middle|\, \mathcal{D}_T \right]}_{\text{(iv)}}
+ \sum_{i=1}^{I}
\underbrace{\mathrm{E}\!\left[ \bigl(\Psi_{\mu}^{(i)}(\theta_t)\bigr)^2 \,\middle|\, \mathcal{D}_T \right]}_{\text{(v)}}
\Big\}
+ C
\Bigg)
\label{eq:Expectedloglikelihood_deriv}
\end{align}

\subsection{Derivation of Taylor Linearization Approach} \label{apx:Taylor}

We consider Taylor expanding $\tau$ at $x_0$ to the first degree:

\begin{align*}
    \tau(x) &= \frac{1}{1+e^{-x}}\\
    &= \frac{1}{1 + e^{-x_0}} + \frac{e^{-x_0}}{(1 + e^{-x_0})^2} (x - x_0) + O((x - x_0)^2)\\
    \eta(x_0) &:= \frac{e^{-x_0}}{(1 + e^{-x_0})^2}
\end{align*}

Take in linearized function evaluated at $\bm{m}_t^*$ into calculation:

\begin{align*}
    \left. \begin{bmatrix}
    \btheta_t\\
    \btheta_{t-1}
    \end{bmatrix} \right\vert \cD_t
    &\sim  \rN \left(
    \begin{bmatrix}
    \bm{m}_t^*\\
    \bm{m}_{t-1}^*
    \end{bmatrix}
    ,
    \begin{bmatrix}
    \bC_t^* & \bm{\Sigma}_{t, t-1}^*\\
    \bm{\Sigma}_{t-1, t}^* & \bC_{t-1}^*
    \end{bmatrix}
    \right)\\
    M &= \bG _{in} \bu_t^2 + \bias\\
    \tau(\bG \btheta_{t-1} + M) &= \tau(\bG \bm{m}_{t-1}^* + M) + \eta(\bG \bm{m}_{t-1}^* + M) \odot \bG  (\btheta_{t-1} - \bm{m}_{t-1}^*) + O((\btheta_{t-1} - \bm{m}_{t-1}^*)^2)\\
    \bE[ \btheta_t^T \bW^{-1} \Xi_\mu( \btheta_{t-1})  |\cD_T] &\approx \mathbb{E}\left[ \btheta_t^T \bW^{-1} \tau(\bG \bm{m}_{t-1}^* + M) \right.\\
    & \quad + \left. (\btheta_t^T-\bm{m}_{t-1}^*)  \bW^{-1} \eta(\bG \bm{m}_{t-1}^* + M) \odot \bG  (\btheta_{t-1} - \bm{m}_{t-1}^*)) \right] \\
    &= (\bm{m}_{t}^*)^T \bW^{-1} \tau(\bG \bm{m}_{t-1}^* + M) \\
    &\quad + \bE((\btheta_t^T-\bm{m}_{t-1}^*) \bW^{-1}\eta(\bG \bm{m}_{t-1}^* + M)\odot \bG  (\btheta_{t-1} - \bm{m}_{t-1}^*))) 
\end{align*}

Taking another substitution,

\begin{align*}
    N &= \eta(\bG \bm{m}_{t-1}^* + M) = (q_1, q_2, \cdots, q_p)^T\\
    \bE[ \btheta_t^T \bW^{-1} \Xi_\mu( \btheta_{t-1})  |\cD_T] &\approx  (\bm{m}_{t}^*)^T \bW^{-1} \tau(\bG \bm{m}_{t-1}^* + M) \\
    &\quad + \bE\left[(\btheta_t^T - (\bm{m}_t^*)^T) \bW^{-1}\left[N\odot \bG  (\btheta_{t-1} - \bm{m}_{t-1}^*)\right]\right]
\end{align*}

Finally, this could be evaluated by standard Multivariate Gaussian Distribution properties.

\begin{align*}
    \left. \begin{bmatrix}
    \bW^{-1}(\btheta_t-\bm{m}_{t}^*)\\
    \bG (\btheta_{t-1}-\bm{m}_{t-1}^*)
    \end{bmatrix} \right\vert \cD_t
    &\sim  \rN \left(
    \begin{bmatrix}
    0\\
    0
    \end{bmatrix}
    ,
    \begin{bmatrix}
    \bW^{-1} \bC_t^* \bW^{-T} & \bW^{-1} \bm{\Sigma}_{t, t-1}^*\bG ^T\\
    \bG \bm{\Sigma}_{t-1, t}^*\bW^{-1} & \bG \bC_{t-1}^* \bG^T
    \end{bmatrix}
    \right)\\
    \bE\left\{(\btheta_t^T - (\bm{m}_t^*)^T) \bW^{-1} \left[N\odot \bG  (\btheta_{t-1} - \bm{m}_{t-1}^*)\right]\right\} &= \sum_{i=1}^p q_i [\bW^{-1}\bm{\Sigma}_{t, t-1}^*\bG ^T]_{[i, i]}
\end{align*}

Therefore, by taking the calculated result back in, we can get the answer

\begin{align*}
    \bE[ \btheta_t^T \bW^{-1} \Xi_\mu( \btheta_{t-1})  |\cD_T] &\approx (\bm{m}_{t}^*)^T \bW^{-1} \tau(\bG \bm{m}_{t-1}^* + M) + \sum_{i=1}^p q_i [\bW^{-1}\bm{\Sigma}_{t, t-1}^*\bG ^T]_{i, i}\\
    &= (\bm{m}_{t}^*)^T \bW^{-1} \tau(\bG \bm{m}_{t-1}^* + M) + N \cdot \mathrm{diag}[\bW^{-1}\bm{\Sigma}_{t, t-1}^*\bG ^T]
\end{align*}

\end{appendices}

\vskip 0.2in
\bibliography{reference}

@inproceedings{Wan2000a,
author = {Wan, Eric A. and {Van Der Merwe}, Rudolph and Nelson, Alex T.},
booktitle = {Advances in Neural Information Processing Systems},
pages = {667--672},
title = {{Dual estimation and the unscented transformation}},
year = {2000}
}

@misc{Manjunath2013,
author = {Manjunath, G. and Jaeger, H.},
booktitle = {Neural Computation},
number = {3},
pages = {671--696},
title = {{Echo state property linked to an input: Exploring a fundamental characteristic of recurrent neural networks}},
volume = {25},
year = {2013}
}

@misc{Wankel2012,
author = {Wankel, Charles and Campbell, Bruce A.},
booktitle = {Encyclopedia of Business in Today's World},
title = {{S{\&}P 500 Index}},
url = {https://finance.yahoo.com/quote/\%5ESPX/history/},
urldate = {2020-05-17},
year = {2012}
}

@article{Sun2017,
author = {Sun, Xiaochuan and Li, Tao and Li, Qun and Huang, Yue and Li, Yingqi},
journal = {Knowledge-Based Systems},
pages = {17--29},
title = {{Deep belief echo-state network and its application to time series prediction}},
volume = {130},
year = {2017}
}

@misc{Jaeger2002,
author = {Jaeger, H},
booktitle = {GMD Report 152},
pages = {60},
title = {{Short term memory in echo state networks}},
year = {2002}
}

@article{Verstraeten2007,
author = {Verstraeten, D. and Schrauwen, B. and D'Haene, M. and Stroobandt, D.},
journal = {Neural Networks},
number = {3},
pages = {391--403},
title = {{An experimental unification of reservoir computing methods}},
volume = {20},
year = {2007}
}

@article{Stoll1969,
author = {Stoll, Hans R.},
journal = {The Journal of Finance},
number = {5},
pages = {801--824},
title = {{the Relationship Between Put and Call Option Prices}},
volume = {24},
year = {1969}
}

@article{Atchison1980,
author = {Atchison, J. and Shen, S. M.},
journal = {Biometrika},
number = {2},
pages = {261--272},
title = {{Logistic-normal distributions: Some properties and uses}},
volume = {67},
year = {1980}
}

@article{Cox1985,
author = {Cox, John C. and Ingersoll, Jonathan E. and Ross, Stephen A.},
journal = {Econometrica},
number = {2},
pages = {385},
title = {{A Theory of the Term Structure of Interest Rates}},
volume = {53},
year = {1985}
}

@article{Bollerslev1986,
author = {Bollerslev, Tim},
journal = {Journal of Econometrics},
number = {3},
pages = {307--327},
title = {{Generalized autoregressive conditional heteroskedasticity}},
volume = {31},
year = {1986}
}

@article{Jaeger2004,
author = {Jaeger, Herbert and Haas, Harald},
journal = {Science},
number = {5667},
pages = {78--80},
title = {{Harnessing Nonlinearity: Predicting Chaotic Systems and Saving Energy in Wireless Communication}},
volume = {304},
year = {2004}
}

@misc{Chouikhi2018,
author = {Chouikhi, Naima and Ammar, Boudour and Alimi, Adel M.},
booktitle = {arXiv},
title = {{Genesis of basic and multi-layer echo state network recurrent autoencoder for efficient data representations}},
year = {2018}
}

@article{West1999,
author = {West, Mike and Prado, Raquel and Krystal, Andrew D. and Prado, Raquel},
journal = {Journal of the American Statistical Association},
number = {446},
pages = {375--387},
title = {{Evaluation and Comparison of EEG Traces: Latent Structure in Nonstationary Time Series}},
volume = {94},
year = {1999}
}

@inproceedings{Doerr2018,
archivePrefix = {arXiv},
arxivId = {1801.10395},
author = {Doerr, Andreas and Daniel, Christian and Schiegg, Martin and Nguyen-Tiong, Duy and Schaal, Stefan and Toussaint, Marc and Trimpe, Sebastian},
booktitle = {35th International Conference on Machine Learning, ICML 2018},
eprint = {1801.10395},
pages = {2060--2075},
title = {{Probabilistic Recurrent State-Space Models}},
volume = {3},
year = {2018}
}

@article{Psiaki2005,
author = {Psiaki, Mark L.},
journal = {Journal of Guidance, Control, and Dynamics},
number = {5},
pages = {885--894},
title = {{Backward-smoothing extended Kalman filter}},
volume = {28},
year = {2005}
}

@article{Black1973,
author = {Black, Fischer and Scholes, Myron},
journal = {Journal of Political Economy},
number = {3},
pages = {637--657},
title = {{The pricing of options and corporate liabilities}},
volume = {81},
year = {1973}
}

@inproceedings{Agamennoni2011,
author = {Agamennoni, Gabriel and Nieto, Juan I. and Nebot, Eduardo M.},
booktitle = {Proceedings - IEEE International Conference on Robotics and Automation},
pages = {1551--1558},
title = {{An outlier-robust Kalman filter}},
year = {2011}
}

@misc{ICEBenchmarkAdministrationLimited2015,
author = {{ICE Benchmark Administration Limited}},
publisher = {FRED, Federal Reserve Bank of St. Louis},
title = {{3-Month London Interbank Offered Rate (LIBOR)}},
url = {https://research.stlouisfed.org/fred2/release?rid=253{\&}t=3-month{\&}ob=pv{\&}od=desc},
year = {2015}
}

@article{Liu1998,
author = {Liu, Jun S. and Chen, Rong},
journal = {Journal of the American Statistical Association},
number = {443},
pages = {1032--1044},
title = {{Sequential monte carlo methods for dynamic systems}},
volume = {93},
year = {1998}
}

@techreport{Jaeger2010,
author = {Jaeger, Herbert},
booktitle = {GMD Report},
institution = {German National Research Center for Information Technology},
title = {{The "echo state" approach to analysing and training recurrent neural networks with an erratum note}},
year = {2010}
}

@article{Lee2014,
archivePrefix = {arXiv},
arxivId = {1206.1623},
author = {Lee, Jason D. and Sun, Yuekai and Saunders, Michael A.},
eprint = {1206.1623},
journal = {SIAM Journal on Optimization},
number = {3},
pages = {1420--1443},
title = {{Proximal Newton-type methods for minimizing composite functions}},
volume = {24},
year = {2014}
}

@article{Julier2004a,
author = {Julier, Simon J. and Uhlmann, Jeffrey K.},
journal = {Proceedings of the IEEE},
number = {3},
pages = {401--422},
title = {{Unscented filtering and nonlinear estimation}},
volume = {92},
year = {2004}
}

@article{Sarkka2008,
author = {S{\"{a}}rkk{\"{a}}, Simo},
journal = {IEEE Transactions on Automatic Control},
number = {3},
pages = {845--849},
title = {{Unscented Rauch-Tung-Striebel smoother}},
volume = {53},
year = {2008}
}

@article{Lukosevicius2012,
author = {Luko{\v{s}}evi{\v{c}}ius, Mantas},
journal = {Lecture Notes in Computer Science (including subseries Lecture Notes in Artificial Intelligence and Lecture Notes in Bioinformatics)},
pages = {659--686},
title = {{A practical guide to applying echo state networks}},
volume = {7700 LECTU},
year = {2012}
}

@article{Heston1993,
author = {Heston, Steven L.},
journal = {Review of Financial Studies},
number = {2},
pages = {327--343},
title = {{A Closed-Form Solution for Options with Stochastic Volatility with Applications to Bond and Currency Options}},
volume = {6},
year = {1993}
}

@article{Kalman1960,
author = {Kalman, R. E.},
journal = {Journal of Fluids Engineering, Transactions of the ASME},
number = {1},
pages = {35--45},
title = {{A new approach to linear filtering and prediction problems}},
volume = {82},
year = {1960}
}

@article{Patriksson1998,
author = {Patriksson, Michael},
journal = {SIAM Journal on Optimization},
number = {2},
pages = {561--582},
title = {{Cost approximation: A unified framework of descent algorithms for nonlinear programs}},
volume = {8},
year = {1998}
}

@article{Crook2007,
author = {Crook, Nigel},
journal = {Neurocomputing},
number = {7-9},
pages = {1167--1176},
title = {{Nonlinear transient computation}},
volume = {70},
year = {2007}
}

@misc{Ma2017,
archivePrefix = {arXiv},
arxivId = {1711.05255},
author = {Ma, Qianli and Shen, Lifeng and Cottrell, Garrison W.},
booktitle = {arXiv},
eprint = {1711.05255},
title = {{Deep-ESN: A multiple projection-encoding hierarchical reservoir computing framework}},
year = {2017}
}

@article{Maass2002,
author = {Maass, Wolfgang and Natschl{\"{a}}ger, Thomas and Markram, Henry},
journal = {Neural Computation},
number = {11},
pages = {2531--2560},
title = {{Real-time computing without stable states: A new framework for neural computation based on perturbations}},
volume = {14},
year = {2002}
}

@article{Tibshirani1996,
author = {Tibshirani, Robert},
journal = {Journal of the Royal Statistical Society: Series B (Methodological)},
number = {1},
pages = {267--288},
title = {{Regression Shrinkage and Selection Via the Lasso}},
volume = {58},
year = {1996}
}

@misc{WhartonSchoolattheUniversityofPennsylvania2013,
author = {{Wharton School at the University of Pennsylvania}},
title = {{Wharton Research Data Services}},
url = {http://wrds-web.wharton.upenn.edu/wrds/},
year = {2013}
}

@article{DelMoral1997,
author = {{Del Moral}, Pierre},
journal = {Comptes Rendus de l'Academie des Sciences - Series I: Mathematics},
number = {6},
pages = {653--658},
title = {{Filtrage non-lin{\'{e}}aire par syst{\`{e}}mes de particules en interaction}},
volume = {325},
year = {1997}
}

@inproceedings{Wan2000,
author = {Wan, E. A. and {Van Der Merwe}, R.},
booktitle = {IEEE 2000 Adaptive Systems for Signal Processing, Communications, and Control Symposium, AS-SPCC 2000},
pages = {153--158},
title = {{The unscented Kalman filter for nonlinear estimation}},
year = {2000}
}

@article{Lukosevicius2009,
author = {Luko{\v{s}}evi{\v{c}}ius, Mantas and Jaeger, Herbert},
journal = {Computer Science Review},
number = {3},
pages = {127--149},
title = {{Reservoir computing approaches to recurrent neural network training}},
volume = {3},
year = {2009}
}

@article{Bertsekas1982,
author = {Bertsekas, Dimitri P.},
journal = {SIAM Journal on Control and Optimization},
number = {2},
pages = {221--246},
title = {{Projected Newton Methods for Optimization Problems with Simple Constraints}},
volume = {20},
year = {1982}
}

@article{Jaeger2007,
author = {Jaeger, Herbert and Luko{\v{s}}evi{\v{c}}ius, Mantas and Popovici, Dan and Siewert, Udo},
journal = {Neural Networks},
number = {3},
pages = {335--352},
title = {{Optimization and applications of echo state networks with leaky- integrator neurons}},
volume = {20},
year = {2007}
}

@article{Jaeger2005,
author = {Jaeger, Herbert},
journal = {ReVision},
pages = {1--46},
title = {{A tutorial on training recurrent neural networks , covering BPPT , RTRL , EKF and the " echo state network " approach}},
volume = {2002},
year = {2005}
}

@misc{JasonAsh2020,
author = {{Jason Ash}},
booktitle = {GitHub repository},
publisher = {GitHub},
title = {pyesg},
url = {https://github.com/jason-ash/pyesg},
year = {2020}
}

@article{Hochreiter1997,
author = {Hochreiter, Sepp and Schmidhuber, J{\"{u}}rgen},
journal = {Neural Computation},
number = {8},
pages = {1735--1780},
title = {{Long Short-Term Memory}},
volume = {9},
year = {1997}
}

@article{Kim2010,
author = {Kim, Dongmin and Sra, Suvrit and Dhillon, Inderjit S.},
journal = {SIAM Journal on Scientific Computing},
number = {6},
pages = {3548--3563},
title = {{Tackling box-constrained optimization via a new projected quasi-newton approach}},
volume = {32},
year = {2010}
}

@article{Maass2000,
author = {Maass, Wolfgang and Sontag, Eduardo D.},
journal = {Neural Computation},
number = {8},
pages = {1743--1772},
title = {{Neural systems as nonlinear filters}},
volume = {12},
year = {2000}
}

@article{Dempster1977,
author = {Dempster, A. P. and Laird, N. M. and Rubin, D. B.},
journal = {Journal of the Royal Statistical Society: Series B (Methodological)},
number = {1},
pages = {1--22},
title = {{Maximum Likelihood from Incomplete Data Via the EM Algorithm}},
volume = {39},
year = {1977}
}

@article{Muller1997,
author = {M{\"{u}}ller, Ulrich A. and Dacorogna, Michel M. and Dav{\'{e}}, Rakhal D. and Olsen, Richard B. and Pictet, Olivier V. and {Von Weizs{\"{a}}cker}, Jacob E.},
journal = {Journal of Empirical Finance},
number = {2-3},
pages = {213--239},
title = {{Volatilities of different time resolutions - Analyzing the dynamics of market components}},
volume = {4},
year = {1997}
}

@article{Pathak2018a,
author = {Pathak, Jaideep and Hunt, Brian and Girvan, Michelle and Lu, Zhixin and Ott, Edward},
journal = {Physical Review Letters},
number = {2},
pmid = {29376715},
title = {{Model-Free Prediction of Large Spatiotemporally Chaotic Systems from Data: A Reservoir Computing Approach}},
volume = {120},
year = {2018}
}

@inproceedings{caonima,
  author={M. D. {Skowronski} and J. G. {Harris}},
  booktitle={2007 IEEE International Symposium on Circuits and Systems}, 
  title={Noise-robust automatic speech recognition using a discriminative echo state network}, 
  year={2007},
  volume={},
  number={},
  pages={1771-1774}}

@misc{Scardapane2017,
author = {Scardapane, Simone and Wang, Dianhui},
booktitle = {Wiley Interdisciplinary Reviews: Data Mining and Knowledge Discovery},
number = {2},
title = {{Randomness in neural networks: an overview}},
volume = {7},
year = {2017}
}

@article{Rodan2017,
author = {Rodan, Ali and Sheta, Alaa F. and Faris, Hossam},
journal = {Soft Computing},
number = {22},
pages = {6811--6824},
title = {{Bidirectional reservoir networks trained using SVM + privileged information for manufacturing process modeling}},
volume = {21},
year = {2017}
}

@article{Bianchi2015,
author = {Bianchi, Filippo Maria and {De Santis}, Enrico and Rizzi, Antonello and Sadeghian, Alireza},
journal = {IEEE Access},
pages = {1931--1943},
title = {{Short-Term Electric Load Forecasting Using Echo State Networks and PCA Decomposition}},
volume = {3},
year = {2015}
}

@article{Bianchi2015a,
author = {Bianchi, Filippo Maria and Scardapane, Simone and Uncini, Aurelio and Rizzi, Antonello and Sadeghian, Alireza},
journal = {Neural Networks},
pages = {204--213},
pmid = {26413714},
title = {{Prediction of telephone calls load using Echo State Network with exogenous variables}},
volume = {71},
year = {2015}
}

@article{Deihimi2012,
author = {Deihimi, Ali and Showkati, Hemen},
journal = {Energy},
number = {1},
pages = {327--340},
title = {{Application of echo state networks in short-term electric load forecasting}},
volume = {39},
year = {2012}
}

@article{Li2012,
author = {Li, Decai and Han, Min and Wang, Jun},
journal = {IEEE Transactions on Neural Networks and Learning Systems},
number = {5},
pages = {787--797},
pmid = {24806127},
title = {{Chaotic time series prediction based on a novel robust echo state network}},
volume = {23},
year = {2012}
}

@article{Shi2007,
author = {Shi, Zhiwei and Han, Min},
journal = {IEEE Transactions on Neural Networks},
number = {2},
pages = {359--372},
pmid = {17385625},
title = {{Support vector echo-state machine for chaotic time-series prediction}},
volume = {18},
year = {2007}
}

@article{Trentin2015,
author = {Trentin, Edmondo and Scherer, Stefan and Schwenker, Friedhelm},
journal = {Pattern Recognition Letters},
pages = {4--12},
title = {{Emotion recognition from speech signals via a probabilistic echo-state network}},
volume = {66},
year = {2015}
}

@article{Hodges1991,
author = {Hodges, James S. and West, Mike and Harrison, Jeff},
journal = {Journal of the American Statistical Association},
number = {414},
pages = {547},
title = {{Bayesian Forecasting and Dynamic Models.}},
volume = {86},
year = {1991}
}

@article{Ljung1979,
author = {Ljung, Lennart},
journal = {IEEE Transactions on Automatic Control},
number = {1},
pages = {36--50},
title = {{Asymptotic Behavior of the Extended Kalman Filter as a Parameter Estimator for Linear Systems}},
volume = {24},
year = {1979}
}

@article{luo2025sequda,
  title={SeqUDA-Rec: Sequential User Behavior Enhanced Recommendation via Global Unsupervised Data Augmentation for Personalized Content Marketing},
  author={Luo, Ruihan and Chen, Xuanjing and Ding, Ziyang},
  journal={Economics and Management Innovation},
  volume={2},
  number={7},
  pages={1--7},
  year={2025}
}

@article{li2025llm,
  title={LLM-based Personalized Portfolio Recommender: Integrating Large Language Models and Reinforcement Learning for Intelligent Investment Strategy Optimization},
  author={Li, Bangyu and Gu, Boping and Ding, Ziyang},
  journal={arXiv preprint arXiv:2512.12922},
  year={2025}
}

\end{document}